\documentclass[sigconf]{acmart}

\AtBeginDocument{\providecommand\BibTeX{{\normalfont B\kern-0.5em{\scshape i\kern-0.25em b}\kern-0.8em\TeX}}}

\copyrightyear{2021}
\acmYear{2021}
\setcopyright{acmcopyright}\acmConference[FAccT '21]{Conference on Fairness,
Accountability, and Transparency}{March 3--10, 2021}{Virtual Event, Canada}
\acmBooktitle{Conference on Fairness, Accountability, and Transparency (FAccT '21),
March 3--10, 2021, Virtual Event, Canada}
\acmPrice{15.00}
\acmDOI{10.1145/3442188.3445865}
\acmISBN{978-1-4503-8309-7/21/03}

\usepackage{hyperref}
\usepackage{subcaption}
\usepackage{algorithm}
\usepackage{algorithmic}
\usepackage{color}
\usepackage{bbm}
\usepackage{enumitem}
\usepackage{multirow}
\usepackage{makecell}
\usepackage{tikz}
\usetikzlibrary{shapes,decorations,arrows,calc,arrows.meta,fit,positioning}
\tikzset{
    -Latex,auto,node distance =0.6 cm and 0.6 cm, thick, line width = 1.5,
    state/.style ={circle, draw, thick, minimum width = 0.8 cm, line width=1pt},
    point/.style = {circle, draw, inner sep=0.04cm,fill,node contents={}},
    bidirected/.style={Latex-Latex,dashed},
    el/.style = {inner sep=2pt, align=left, sloped},
    every picture/.style={line width=1pt}
}

\DeclareMathOperator*{\argmin}{arg\,min}

\global\long\def\reals{\mathbf{R}}
\global\long\def\bfX{\mathbf{X}}
\global\long\def\bfS{\mathbf{S}}

\global\long\def\bfx{\mathbf{x}}

\global\long\def\bfs{\mathbf{s}}

\global\long\def\yh{\smash{\hat{Y}}}
\global\long\def\pr{\text{Pr}}
\global\long\def\pd{P}
\global\long\def\ep{\mathbb{E}}
\global\long\def\cn{\mathcal{N}}
\global\long\def\ind{\mathbbm{1}}

\global\long\def\std{\texttt{Standard}}
\global\long\def\cda{\texttt{CausalDA}}
\global\long\def\stdf{\texttt{Standard+FairLearn}}
\global\long\def\cdaf{\texttt{CausalDA+FairLearn}}
\global\long\def\ot{\texttt{OTDA}}
\global\long\def\fl{\texttt{FairLearn}}
\global\long\def\ar{\texttt{AnchorReg}}

\newcommand\independent{\protect\mathpalette{\protect\independenT}{\perp}}
\def\independenT#1#2{\mathrel{\rlap{$#1#2$}\mkern2mu{#1#2}}}

\providecommand{\tightlist}{\setlength{\itemsep}{0pt}\setlength{\parskip}{0pt}}
  
\newcommand{\specialcell}[2][c]{\begin{tabular}[#1]{@{}c@{}}#2\end{tabular}}

\newtheorem{prop}{\textbf{Proposition}}
\newtheorem{assumption}{\textbf{Assumption}}
\newtheorem{definition}{\textbf{Definition}}[section]
\newtheorem{lemma}{\textbf{Lemma}}
\newtheorem{theorem}{\textbf{Theorem}}
\theoremstyle{remark}
\newtheorem{remark}{Remark}

\begin{document}

\title{Fairness Violations and Mitigation under Covariate Shift}

\author{Harvineet Singh}
\affiliation{\institution{Center for Data Science\\New York University}
  \city{New York City}
  \state{NY}
  \country{USA}
}
\email{hs3673@nyu.edu}

\author{Rina Singh}
\authornote{Work done while at New York University. Current affiliation is Fusemachines Inc.}
\affiliation{\institution{Tandon School of Engineering\\New York University}
  \city{New York City}
  \state{NY}
  \country{USA}
}
\email{rina@fusemachines.com}

\author{Vishwali Mhasawade}
\affiliation{\institution{Tandon School of Engineering\\New York University}
  \city{New York City}
  \state{NY}
  \country{USA}
}
\email{vishwalim@nyu.edu}

\author{Rumi Chunara}
\affiliation{\institution{Tandon School of Engineering;\\School of Global Public Health\\New York University}
  \city{New York City}
  \state{NY}
\country{USA}
}
\email{rumi.chunara@nyu.edu}

\renewcommand{\shortauthors}{Singh, et al.}

\begin{abstract}
  We study the problem of learning fair prediction models for unseen test sets distributed differently from the train set. Stability against changes in data distribution is an important mandate for responsible deployment of models. The domain adaptation literature addresses this concern, albeit with the notion of stability limited to that of prediction accuracy. We identify sufficient conditions under which stable models, both in terms of prediction accuracy and fairness, can be learned. Using the causal graph describing the data and the anticipated shifts, we specify an approach based on feature selection that exploits conditional independencies in the data to estimate accuracy and fairness metrics for the test set. We show that for specific fairness definitions, the resulting model satisfies a form of worst-case optimality. In context of a healthcare task, we illustrate the advantages of the approach in making more equitable decisions.
\end{abstract}

\ccsdesc[500]{Computing methodologies~Learning under covariate shift}
\ccsdesc[500]{Computing methodologies~Causal reasoning and diagnostics}

\keywords{algorithmic fairness, domain adaptation, covariate shift, causal inference}

\maketitle

\section{Introduction}

Deployment of machine learning algorithms to aid consequential decisions, such as in medicine, criminal justice, and employment, require revisiting the dominant paradigms of training and testing such algorithms. Particularly, the assumption that the data distribution in training and deployment will be the same is not always warranted.
Examples of the impact of distribution shift can be found in medical imaging tasks \citep{zech2018variable,pooch2019can}, where the algorithms trained on one chest radiography dataset performed poorly on other datasets. Similarly, \citet{nestor2019feature} find that models for critical care tasks degraded in performance over time resulting from changes in the instrumentation of the electronic health records.
Given the safety-critical nature of the decisions, the decision-making process should account for these shifts in distributions to ensure high predictive accuracy of the algorithms.

Many methods exist to learn under distribution shifts \citep{quionero2009dataset}, including recent work from a causal inference perspective \citep{rojas2018invariant, subbaswamy2019preventing, pearl2011transportability, arjovsky2019invariant}. Such methods have significant appeal since they allow learning accurate models for \textit{arbitrary} shifts, including those in unseen future data. This is achieved by exploiting causally-relevant factors in data that are generalizable to unseen test sets, as opposed to fitting to the factors specific to the training sets. However, the focus of the methods has been on \textit{average} case prediction performance alone.
In certain circumstances, while high predictive accuracy is a necessary requirement, decisions made using the algorithms should also not lead to or perpetuate past disparities among \textit{groups} in the data. 
Without any design changes, algorithmic solutions for mitigating distribution shifts that do not account for disparities in training data can result in disparate impact while predicting under distribution shifts. We discuss a concrete example later.

At the same time, most work in algorithmic fairness addresses the setting with a \textit{single} learning task (or domain) under the assumption that the data distribution does not change between train and test settings \citep{hardt2016equality, agarwal2018reductions}. Under this assumption, minimizing classification risk along with constraints on the fairness metric in training data is likely to generalize to identically distributed test data. 
Thinking about shifts in fair machine learning is also important though, since deployment of a (fair) decision-making tool might affect what data is collected in future (e.g. selectively policing locations with high predicted risk \citep{lum2016predict}), or might incentivize individuals to strategically adapt their features for favourable outcomes \citep{hardt2016strategic, liu2019delayed}, thus, causing distribution shift. In addition, due to data-scarcity, such as in medical decision-making \citep{wiens2014study}, the models may be applied to newer settings (such as hospitals) than the ones seen during training. The issue of ensuring fairness when deployment environment differs from the training one has received little attention \citep{schumann2019transfer}. Due to the variety of train-test shifts that can occur, conceptualizing and addressing the problem has been challenging.

\textbf{Our contributions.} We address the problem of learning fair models under mismatch in train-test distributions when either limited or no data is available from the test distribution.
We consider the setup of \textit{causal domain adaptation} where possible shifts are expressed using causal graphs with the goal of learning models with stable performance under the specified shifts. Our main contribution is to formulate the fair learning problem in this setup and provide sufficient conditions that enable estimation of model accuracy and fairness metrics in the test domain. For a subset of covariate shifts and for several well-known group-fairness metrics, we show that the resulting solution is worst-case optimal. We operationalize the sufficient conditions in an algorithm based on a reduction to the standard fair learning problem. Finally, we present a case study on a medical decision-making task which demonstrates applicability of the approach.

\section{Related work}

Domain adaptation and fair machine learning are both widely studied problems. Thus, we primarily focus the discussion on literature at their intersection.

\textbf{Fairness.}
A number of fairness metrics have been proposed that make different normative statements on the machine learning models' output (see \citep{mitchell2018prediction} for a review).
Depending on the application context, different metrics might be appropriate or mandatory by law \citep{narayanan2018translation}.
Consequently, \textit{fairness methods} have been developed to build/modify models that satisfy different fairness criteria.
We focus on a class of methods that pose the problem as that of constrained optimization \citep{agarwal2018reductions,donini2018empirical}.

\textbf{Domain adaptation.}
The seminal work of \citet{ben2010theory} relates the target domain error to the source domain error and the distance between the distributions. This inspired many domain adaptation methods based on adversarial training of representations to align the distributions \citep{ganin2016domain}. One drawback is that the methods require some data from test distribution while training.
When causal structure of the domains is known, recent work on \textit{causal domain adaptation} \citep{rojas2018invariant, magliacane2018domain, subbaswamy2019preventing,  subbaswamy2018counterfactual} identify predictors with stable accuracy under unseen changes in distribution.
To accomplish this, the methods exploit the principle of invariance of causal mechanisms \citep[][Sec. 1.3]{pearl2009causality} that says -- interventions (or shifts) in certain mechanisms in the graph keep the other mechanisms unchanged. The invariant mechanisms can be used to build stable predictors.
Similar to \citep{subbaswamy2019preventing, pearl2011transportability}, we adopt a setting where a causal graph specifies anticipated distribution shifts and no target domain data is given (but can be used if available). The goal is to construct predictors that are invariant to all anticipated shifts, without necessarily observing the corresponding data. The setting is particularly well-suited for consequential decision-making where we want to \textit{proactively} guard against shifts that may result in harm, before deploying the model and collecting target data.
However, none of these methods consider the possibility of unfair outcomes after adaptation.

\textbf{Fairness and domain adaptation.}
On multiple benchmark datasets, \citet{friedler2019comparative} found that fair machine learning methods showed high variance in achieved accuracy and fairness on randomly split train-test sets. To mitigate this, \citet{huang2019stable} propose adding a regularization term to the constrained ERM problem that guarantees \textit{stability}. However, the term stability is used for changes in the fairness metric as a training data sample is removed/added, as opposed to changes under different distributions. 
In \citep{cotter2019training}, authors propose algorithms for generalisation of fairness constraints but to an i.i.d. test set.
In \citep{madras2018learning}, the authors propose learning feature representations, using adversarial training, which result in fair classifiers when trained on the representations. They do not address changes in distribution of the features (and their representations) across domains.

In the same setup as ours under the assumption of covariate shift but with the availability of unlabelled target data, \citep{coston2019fair} give weighting-based estimators and \citep{rezaei2020robust} take a robust optimization approach.
Other works that assume some labelled data from the target domain include \citep{oneto2019learning,schumann2019transfer, slack2020fairness}. 
For instance, \citep{oneto2019learning} learns a representation from multiple domains with guarantees on generalization to the target domain, but requires labelled target data to fine-tune classifiers and a low-rank assumption that constrains dis-similarity between the domains.
In \citep{slack2020fairness}, authors restrict to shifts in feature means and propose ways to flag a potentially unfair model under such shifts. Further, concurrent work \citep{mandal2020ensuring} posits a set of test distributions defined as weighted combinations of the training data, and find a fair classifier minimizing the worst loss across such distributions.
Instead, we rely on distributional assumptions expressed using a causal graph.
Considering the causal structure of the problem allows the modeller to express plausible distribution shifts more intuitively by denoting the mechanisms, instead of the statistical properties, that can change. It also guides the construction of estimators that are robust against shifts of \textit{arbitrary} magnitude rather than only the shifts in the observed datasets.

Our work is related in spirit to \citep{blum2020recovering, kallus2018residual} who consider building fair models from `biased' training data. Here, we provide a complementary set of results on fairness under train-test distribution mismatch, avoiding assumptions on specific generative processes for the shift. Instead we use causal graphs to make weaker assumptions on where the mismatch is. This allows us to give a general characterization of the addressable mismatch settings. Moreover, at a conceptual level, our focus is on addressing mismatch with multiple \textit{future} test sets rather than a biased training set.

\section{Problem setup}

Let us denote all the variables associated with the system being modelled as $\textbf{V} := \left(\textbf{X},A,Y\right)$, where $A$ is the sensitive attribute, $\textbf{X}$ is a non-empty set of covariates other than $A$, and $Y$ is an outcome of interest.
We will consider a binary sensitive attribute, $A\in \{\texttt{a},\texttt{d}\}$ (i.e. \textit{advantaged} and \textit{disadvantaged} group), and the binary classification case, thus, $\smash{Y\in \{0,1\}}$. 
For simplicity of exposition, consider the case with only two domains -- a source and a target -- with joint probability distributions $\pd_\text{source}$ and $\pd_\text{target}$, respectively. Crucially, the two distributions may be different (e.g. data from two hospitals with different care practices).
Bold letters are used for vectors, uppercase for random variables, and lowercase for instantiations.

\subsection{Fair classifier}
\label{sec:fair_pred}
Consider that the classifier is built from the feature (sub)set $\smash{\bfS \subseteq \{\textbf{X},A\}}$ and outputs the binary prediction $f(\bfS)\in \{0,1\}$.\footnote{Note that $\bfS$ can contain $A$ as we assume that disparate treatment is allowed in the problems of our interest.} 
We will operate in the empirical risk minimization framework for learning classifiers and introduce additional fairness constraints in the objective to control the inter-group disparity, a commonly-used approach \cite{donini2018empirical, agarwal2018reductions, zafar2019fairness}.
Each constraint is given by some function $G$ of the prediction, outcome, and features. Denote the constraint by $G(f(\bfS),(Y,\bfS))\leq \epsilon$ with $(Y,\bfS)\sim \pd_\text{target}(Y,\bfS)$ and some hyperparameter $\epsilon\geq 0$ allowing for \textit{approximate} fairness.
If there are multiple constraints, we write the set of constraints succinctly as $\textbf{G}(f, \pd_\text{target})\leq \boldsymbol{\epsilon}$.
Note that the desired fairness constraint $\textbf{G}$ is assumed to be the same in both the domains. 
The classification error i.e. probability of a misclassification is written as $\pd(f(\bfS)\neq Y)$.
Then, the fair domain adaptation (DA) problem amounts to finding a minimizer
\begin{align}
\label{eq:ferm}
    \text{[Fair DA]}\ \ {f^*_\text{target}} := \argmin_{f \in \mathcal{F}(\bfS)}\  \{\pd_\text{target}(f(\bfS)\neq Y) : \textbf{G}(f, \pd_\text{target})\leq \boldsymbol{\epsilon}\}
\end{align}
i.e. a function ${f^*_\text{target}}$ in the set of learnable functions $\mathcal{F}(\bfS)$ of features $\bfS$ that minimizes classification error as well as satisfies fairness constraints. 

\begin{figure*}[t]
\centering
\begin{subfigure}[b]{.33\linewidth}
  \centering
\scalebox{0.8}{\begin{tikzpicture}
\node[state] (a) at (0,0) {$A$};
   
    \node[state,rectangle,fill=blue!10] (c) [below =of a] {$C$}; \node[state] (t) [right =of c] {$T$}; \node[state] (y) [right =of a] {$Y$}; 
    \node[state] (r) [above =of y]{$R$};

    \path (c) edge (t);
    \path (c) edge (a);
    \path (a) edge (t);
    \path (a) edge (y);
    \path (a) edge (r);
    \path (y) edge (t);
    \path (r) edge[bend left=40] (t);
    \path (r) edge  (y);

\end{tikzpicture} }
  \caption{Example causal graph annotated to show anticipated shifts in the distributions $P(A)$ and/or $P(T\vert A,Y)$.}
  \label{fig:synth_data_graph}
\end{subfigure}\hfill
\begin{subfigure}[b]{.33\linewidth}
  \centering
  \includegraphics[scale=0.33]{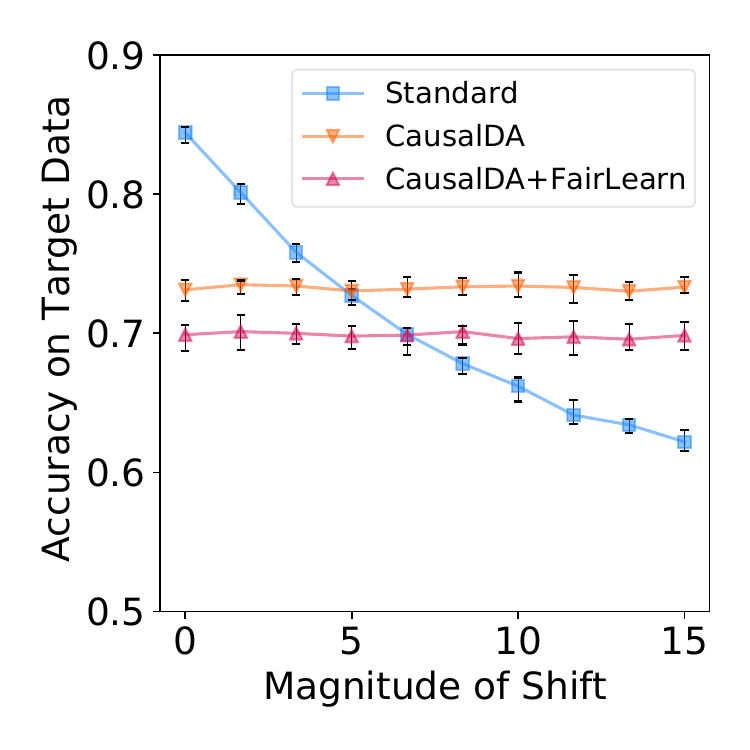}
  \caption{Accuracy vs. Shift.}
  \label{fig:multiple_shift_acc}
\end{subfigure}\hfill
\begin{subfigure}[b]{.33\linewidth}
  \centering
  \includegraphics[scale=0.33]{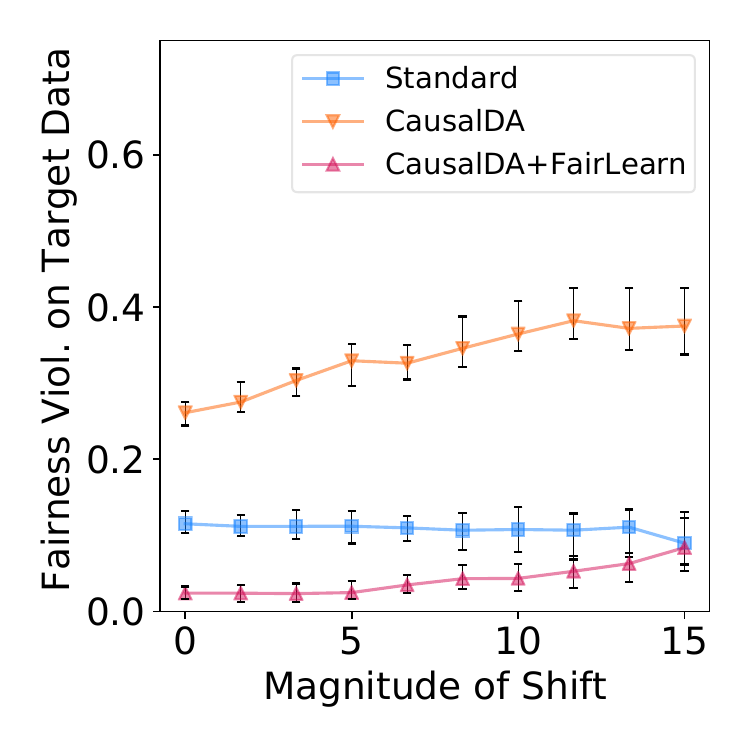}
  \caption{Fairness Violation vs. Shift.}
  \label{fig:multiple_shift_fair}
\end{subfigure}
\caption{Flu diagnosis example. (a) Data generating process for source and target domains represented as a causal graph where domains are indicated by the context variable $C$. Edges from $C$ represent shifts between the domains. $\{T,R,A\}$ are features, with sensitive attribute $A$, and outcome $Y$.
(b,c) Classification accuracy and fairness violation with varying magnitude of shifts for synthetic data (Section \ref{sec:synth_exp}) for the example. Fairness violation is computed as the maximum violation of equalized odds constraint across $Y$ and $A$. Median values are plotted over $50$ runs and error bars show first and third quartiles. Proposed approach ($\cdaf$) achieves both stable accuracy and fairness in the shifted target domains.}
\label{fig:exp_graphs}
\end{figure*}
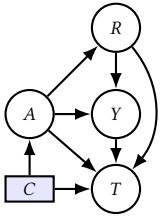
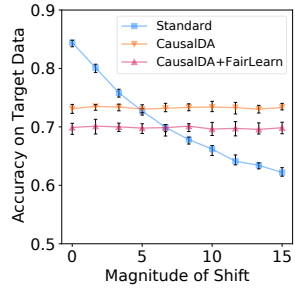
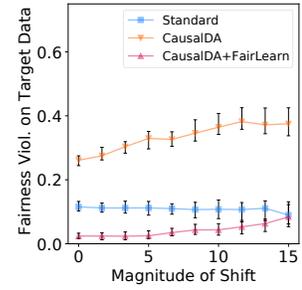

\textbf{Fairness metrics.}
We will focus on group-fairness metrics defined based on some notion of parity across groups. These have received much attention in the fair machine learning literature \citep{agarwal2018reductions, hardt2016equality, donini2018empirical} due to the relative ease of communicating their implications to stakeholders and the ease of computing them from observational data.

\theoremstyle{definition}
\begin{definition}{(DP) \citep{calders2009building}}
A classifier $f$ is said to satisfy \textit{demographic parity} for some distribution $\pd$ if $\pd(f(\bfS)\vert A) = \pd(f(\bfS))$. Thus, the constraint $G$ is $\vert\pd(f(\bfS)\vert A=\texttt{a}) - \pd(f(\bfS)\vert A=\texttt{d})\vert\leq\epsilon$.
\end{definition}

\theoremstyle{definition}
\begin{definition}{(EO) \citep{hardt2016equality}}
A classifier $f$ is said to satisfy \textit{equalized odds} for some distribution $\pd$ if $\pd(f(\bfS)\vert Y=y, A) = \pd(f(\bfS)\vert Y=y)$ for $y\in \{0,1\}$. Thus, the constraints $\textbf{G}$ are $\vert\pd(f(\bfS)\vert Y=y, A=\texttt{a}) - \pd(f(\bfS)\vert Y=y, A=\texttt{d})\vert\leq\epsilon$ for $y\in \{0,1\}$.
\end{definition}

We define two more metrics derived from EO. If we condition only on $Y=1$, the resulting metric is known as \textit{true positive rate equality} (TPR), or more commonly \textit{equality of opportunity} \citep{hardt2016equality}. Similarly, for $Y=0$, the metric is known as \textit{true negative rate equality} (TNR).

Solving (\ref{eq:ferm}) requires estimating the error $\pd_\text{target}(f(\bfS)\neq Y)$ and the fairness constraint $\textbf{G}(f, \pd_\text{target})$.
Given enough samples from $\pd_\text{target}$, standard fair learning methods e.g. \citep{agarwal2018reductions} return a solution. But, this is not possible in the Fair DA setting, as we do not have access to the complete target data. Thus, the central question we ask is:
\textbf{Under what assumptions can we still find ${f^*_\text{target}}$?}
For arbitrary distribution shifts, it is not possible to answer this question in affirmative. With background knowledge of how the distributions differ, past work provides methods to bound the target domain error. Crucially, such methods still do not guarantee target domain fairness and using fairness constraints from the source domain, naturally, does not solve (\ref{eq:ferm}).
Through the following example, we illustrate that these design choices can significantly affect accuracy and fairness of the models.
It also shows how the causal inference framework for domain adaptation allows for the specification of shifts and design of predictors.

\subsection{An illustrative example.}
Consider a simplified version of the flu diagnosis task from \citep{mhasawade2020population}. The associated data generating process is shown in Figure~\ref{fig:synth_data_graph}. Flu status $Y$ of a person is to be predicted from three measurements $\{T,R,A\}$. The disease has two known causes $R$ and $A$, say virus-exposure risk and age group (indicating adult or child) respectively. In addition, a noisy yet predictive symptom of flu is observed as $T$, say body temperature, which is expressed differently depending on the age group. A categorical variable $C$ indicates different data collection sites (the domains) which differ on (i) how well the temperature is measured, e.g. self-reported vs. clinician-tested ($C \rightarrow T$), and (ii) the proportion of demographics across sites ($C \rightarrow A$). 
Suppose, a classifier $\yh$ is to be built using data from a single site (source domain) and used in multiple sites (target domains) to allocate scarce healthcare resources (testing kits, medical consultation) to individuals. The model designer would like to mitigate differential error rates across age groups and chooses to use EO as the fairness constraint while learning $\yh$.

We compare three ways of designing the model that account differently for the possibility of shift and unfairness. 
Figures \ref{fig:multiple_shift_acc}, \ref{fig:multiple_shift_fair} show results from a simulation, discussed in detail in Section \ref{sec:synth_exp}. As we vary the magnitude of distribution shift between the sites, the \texttt{Standard} classifier, built by regressing $Y$ on $\{T,R,A\}$ from source data degrades in accuracy (blue curve) on target data. By accounting for the shift, $\cda$, a domain adaptation approach \citep{rojas2018invariant} that only uses the features $\{R,A\}$, remains stable (orange curve). Surprisingly, domain adaptation leads to higher levels of fairness violations, as shown in Figure \ref{fig:multiple_shift_fair}. To mitigate this we would want to learn $\cda$ with fairness constraints which is complicated, as discussed earlier, since we cannot evaluate the constraints for unseen target domains. However, following the method proposed in Section \ref{sec:fair_dom}, learning $\cda$ with fairness constraints on the \textit{source} domain (red curve) retains both the desired properties -- consistently high accuracy and low unfairness.
Thus, the example illustrates the need to consider fairness constraints while adapting for the shifts.

Next, we describe the joint causal graphs in more detail that allow us to represent the potential shifts, followed by our main results on learning fair and stable predictors under specific shifts.

\section{Joint causal inference and domain adaptation}
\label{sec:causal_dom}

Following recent work \citep{magliacane2018domain,mooij2016joint,rojas2018invariant}, we consider a joint causal graph which represents the data distribution for all domains. 
This allows us to reason about the \textit{invariant} distributions under shifts, which is key to addressing the fair domain adaptation problem.

Assume that all the source and the target domains are characterized by a set of variables $\textbf{V}$, which are observed under different \textit{contexts} (e.g. experimental settings) particular to each domain. Joint Causal Inference \citep[Sec. 3]{mooij2016joint} framework provides a way of representing the data generating process for all domains as a single causal graph representing an underlying causal model. In addition to the \textit{system variables} $\textbf{V}$, the framework introduces an additional set of exogenous variables, named \textit{context variables} $\textbf{C}$ , that represent the modeler's knowledge of how the domains differ from one another (given by the causal relations among the system and context variables).\footnote{In a related concept, selection diagrams also add auxiliary variables to a causal graph to represent the distributions that can change across different domains \citep{pearl2011transportability}. More discussion on the relationship between the two can be found in \cite{mooij2016joint}.} We include the formal definition of JCI framework in Appendix 
\ref{app:jci} along with the necessary assumptions on faithfulness, and Markov property.
For the example in Figure~\ref{fig:synth_data_graph}, system variables are $\{T,R,A,Y\}$. With a binary context variable $C$, $\smash{\pd(T,R,A,Y\mid C=0)}$ and $\smash{\pd(T,R,A,Y\mid C=1)}$ correspond to joint distributions for the two domains, source and target.
More generally, setting context variable to a particular value, say $\textbf{C}=\textbf{c}$, can be seen as an intervention that results in the data distribution for a domain $\smash{\pd(\textbf{V}\mid \textbf{C}=\textbf{c})}$.\footnote{Under the assumptions of JCI framework, discussed in Appendix 
\ref{app:jci}, this is the same as $\smash{\pd(\textbf{V}_{do(\textbf{C}=\textbf{c})})}$ where $do(\textbf{C}=\textbf{c})$ denotes an intervention on $C$.}

A class of causal domain adaptation problems is to learn a predictor that generalizes to different target data distributions which correspond to different settings of the context variables in the causal graph. In \citep{magliacane2018domain}, authors propose learning a predictor using only a \textit{subset} of the features that guarantee invariance of the outcome distribution conditional on the chosen feature subset. More specifically, if $\textbf{V}=(\textbf{X},A,Y)$ and $\textbf{C}$ are the context variables, the desired subset of features $\bfS\subseteq \{\textbf{X},A\}$ satisfies $Y \independent \textbf{C} \mid \bfS$, implying that the conditional distribution of outcome $Y$ given the features $\bfS$ is invariant to the effect of domain changes.
The set $\bfS$ is referred to as a \textit{separating set} as it \textit{d}-separates $Y$ and $\textbf{C}$ in the joint causal graph.
This criterion generalizes the \textit{covariate shift} criterion \citep{sugiyama2008direct}, which assumes independence between $Y$ and $C$ conditioned on all the features.
Note that the separating set criterion excludes graphs where $\textbf{C}$ directly causes $Y$, known as \textit{label shift}.
The predictor using the separating set satisfies a desirable optimality property.
As shown in \citep[]{rojas2018invariant}, it has the lowest mean squared loss against any distribution having the same outcome distribution $Y \mid \bfS$ as in the source.

However, using a separable set in itself does not guarantee fairness. For example, separating sets for Figure \ref{fig:synth_data_graph} are $\bfS\in \{\{A\},\{A,R\}\}$. But neither satisfies the condition required for EO, in general, i.e. $f(\bfS) \independent A\mid Y$. 
Thus, to ensure both invariance and fairness, we restrict our search space in Fair DA (\ref{eq:ferm}) to $\mathcal{F}(\bfS)$, i.e. the set of predictors built using the separating set $\bfS$.
Next, we describe the assumptions that allow us to solve this problem. All proofs are included in Appendices
\ref{app:example}${-}$\ref{app:worstcase} in the supplemental material.

\section{Fair domain adaptation}
\label{sec:fair_dom}

Now, we return to our problem of finding fair classifiers for the target domain and describe how the joint causal graph helps in solving (\ref{eq:ferm}). In the context variable notation, we are interested in finding
$$\argmin_{f \in \mathcal{F}(\bfS)}\  \{\pd(f(\bfS)\neq Y\vert C=1) : \textbf{G}(f, \pd(Y,\bfS\vert C=1))\leq \boldsymbol{\epsilon}\}$$
where $C=1$ represents the target domain. We start by noting the need for further assumptions.

\begin{prop}
\label{prop:example}
Fair DA problem (\ref{eq:ferm}) is not solvable in general without further assumptions.
\end{prop}

Proposition (\ref{prop:example}) follows by the impossibility results on domain adaptation \citep{ben2010impossibility}. Even when domain adaptation is possible, i.e. target domain error is identifiable (uniquely estimable in terms of source domain distribution), the fairness constraint is not guaranteed to be identifiable. We make this point by constructing an example with group-specific measurement error in features.

Thus, the natural question is under what conditions on distributions and assumptions on data availability can we identify the error $\smash{\pd(f(\bfS)\neq Y\vert C=1)}$ and the fairness constraint $\smash{\textbf{G}(f, \pd(Y,\bfS\vert C=1))}$.
We make the following two assumptions for the selected features $\bfS\subseteq \{\textbf{X},A\}$ for the classifier.

\begin{assumption}[Invariance of classification error]
\label{par:assum1} Features $\bfS$ form a separating set, i.e. $C \independent Y \mid \bfS$.
\end{assumption}

\begin{assumption}[Invariance of fairness constraint]
\label{par:assum2}
Depending on the fairness metric, assume that
\begin{itemize}
\item For demographic parity (DP), $\bfS$ satisfies $C \independent \bfS\mid A$,
    \item For equalized odds (EO), $\bfS$ satisfies $C \independent \bfS\mid Y, A$,
    \item For true positive rate equality (TPR), $\bfS$ satisfies $C \independent \bfS\mid Y=1, A$,
    \item For true negative rate equality (TNR), $\bfS$ satisfies $C \independent \bfS\mid Y=0, A$.
\end{itemize}
\end{assumption}

For example, the condition for DP asserts that the characteristics (in terms of features $\bfS$) of the sensitive groups are invariant across domains. Similarly, the condition for EO says that feature distribution for groups defined by the label and the sensitive attribute is invariant across domains. This ensures that we can evaluate (and hence balance) the corresponding fairness constraint irrespective of the domain.

Next, we consider two scenarios to state the quality of the solution that can be found under the two assumptions -- (i) when labelled source and unlabelled target domain data is available, alternatively, (ii) when only the labelled source domain data is available.

\subsection{Fair domain adaptation with limited target domain data}

\begin{prop}
\label{prop:ferm_solution}
Given Assumptions \ref{par:assum1} and \ref{par:assum2} hold, then using only labelled source and unlabelled target data, the Fair DA problem (\ref{eq:ferm}) can be solved exactly by a data re-weighting method.
\end{prop}
\begin{proof}[\textbf{Proof sketch}]
This follows since the error is invariant, i.e. $\pd(f(\bfS)\neq Y\vert \bfS, C=1)=\pd(f(\bfS)\neq Y\vert \bfS, C=0)$, due to Assumption \ref{par:assum1}. This implies that
$$\ep_{Y,\bfS}(\pd(f(\bfS)\neq Y\vert \bfS, C=1))=\ep_{Y,\bfS}(w(\bfS)\times\pd(f(\bfS)\neq Y\vert \bfS, C=0))$$
where weights, $w(\bfS)=\pd(\bfS\vert C=1)/\pd(\bfS\vert C=0)$, are the ratio of feature densities. Under Assumption \ref{par:assum2}, the fairness constraint is invariant, i.e. $\smash{\textbf{G}(f, \pd(Y,\bfS\vert C=1))=\textbf{G}(f, \pd(Y,\bfS\vert C=0))}$.
To solve (\ref{eq:ferm}), we find 
$$\argmin_{f \in \mathcal{F}(\bfS)}\  \left\{w(\bfS)\pd(f(\bfS)\neq Y\vert C=0) : \textbf{G}(f, \pd(Y,\bfS\vert C=0))\leq \boldsymbol{\epsilon}\right\}.$$
Both the error and the constraint are estimable as we have labelled source data sampled from $\pd(Y,\bfS\vert C=0)$.
The remaining term is the density ratio $w(\bfS)$ used to re-weight the error. Since we have features from both source and target in this scenario, $w(\bfS)$ can be computed, for instance, using a probabilistic classifier for discriminating between the domains \citep{bickel2007discriminative}.
\end{proof}
This solution strategy is akin to the importance-weighting approach of addressing covariate shift \citep{shimodaira2000improving, sugiyama2008direct}, with the distinction being the use of the separating feature set instead of all the features.

\begin{figure}[t]
    \begin{subfigure}[b]{.5\linewidth}
\parbox{.6\linewidth}{
          \centering
          \scalebox{0.7}{\begin{tikzpicture}
    \node[state] (a) at (0,0)  {$A$};
    \node[state] (r) [above right =of a] {$R$};
    \node[state] (y) [below right=of r] {$Y$}; 
    \node[state] (l) [below right =of a] {$L$};
    \node[state] (t) [left =of l] {$T$};
     \node[state,rectangle,fill=blue!10] (c2) [left=of t] {$C_2$};
     \node[state, rectangle, fill=blue!10] (c1) [left=of a] {$C_1$};
    \path [color=magenta] (a) edge (y);
    \path (a) edge (r);
    \path [color=green] (y) edge (l);
    \path (r) edge (l);
    \path [color=green] (t) edge (l);
\path [color=green] (c2) edge (t);
    \path [color=magenta] (c1) edge (a);
    \path (a) edge (l);
    \path (r) edge (y);

\end{tikzpicture} }
        }\hfill
        \parbox{.6\linewidth}{
            \centering
            \begin{tabular}{ |c|c| }
                 \hline
                 A & \specialcell{Demography\\(sensitive attribute)} \\
                 \hline
                 Y & Disease status \\ 
                 \hline
                 R & Risk factors\\ 
                 \hline
                 L & Lab tests\\ 
                 \hline
                 T & Treatment\\
                 \hline
            \end{tabular}
        }
      \caption{}
\label{fig:signature_a}
    \end{subfigure}\hfill
    \begin{subfigure}[b]{.5\linewidth}
\parbox{.6\linewidth}{
        \centering
        \scalebox{0.7}{

\begin{tikzpicture}
    \node[state] (a) at (0,0)  {$A$};
    \node[state] (s) [right =of a] {$S$}; 
    \node[state] (l) [above =of s] {$L$};
    \node[state] (y) [right =of s] {$Y$};
    \node[state] (r) [below right =of a] {$R$};

     \node[state,rectangle,fill=blue!10] (c) [left =of a] {$C$}; 
\path (l) edge (y);
    \path (l) edge (s);
    \path (l) edge[bend right] (r);
    \path [color=magenta] (a) edge (s);
    \path (a) edge (r);
    \path (r) edge (y);
    \path [color=magenta] (s) edge (y);
\path [color=magenta] (c) edge (a);

\end{tikzpicture} }}
\hfill
        \parbox{.6\linewidth}{
            \centering
            \begin{tabular}{ |c|c| }
\hline
                 A & Gender (sensitive attribute) \\
                 \hline
                 L & Age \\
                 \hline
                 Y & Credit risk level \\ 
                 \hline
                 R & \specialcell{Repayment duration,\\ Credit amount}\\ 
                 \hline
                 S & Savings, Housing\\
                 \hline
            \end{tabular}
        }
    \caption{}
\label{fig:signature_b}
    \end{subfigure}\caption{Examples of addressable causal graphs. (a) Disease risk scoring under population shift and treatment policy shift \citep{subbaswamy2018counterfactual} (b) Credit scoring under population shift \citep{chiappa2019path}. Following Assumptions \ref{par:assum1} and \ref{par:assum2}, including $A$ in the feature set blocks the effect of population shift (e.g. the paths in magenta) and excluding $L$ from the feature set blocks the effect of treatment policy shift (e.g. the path in green).}
\label{fig:signature}
\end{figure}
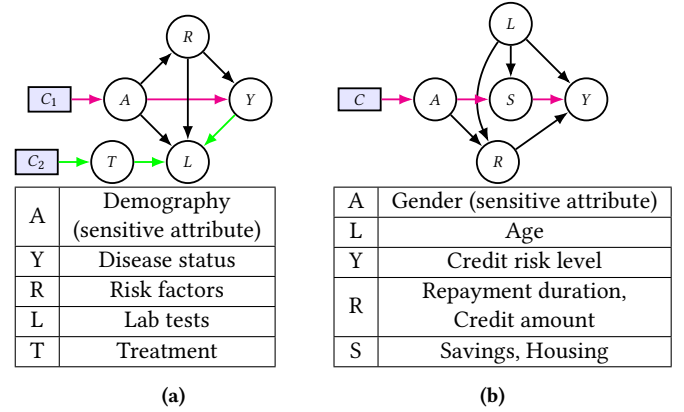

\subsection{Fair domain adaptation with no target domain data}

In the scenario when only the labelled source data is available, we cannot use Proposition (\ref{prop:ferm_solution}) since we cannot estimate the weights. Instead, we use the source data with the selected features,
\begin{equation}
\label{eq:fairda_sol}
\begin{aligned}
    \tilde{f}^*\in\argmin_{f \in \mathcal{F}(\bfS)}\  &\left\{\pd(f(\bfS)\neq Y\vert C=0) : \textbf{G}(f, \pd(Y,\bfS\vert C=0))\leq \boldsymbol{\epsilon}\right\},\\
    &\text{with $\bfS$ satisfying Assumptions \ref{par:assum1} and \ref{par:assum2}}
\end{aligned}
\end{equation}

Next, we show that this solution minimizes the worst-case error under fairness constraints among target distributions satisfying the two assumptions with respect to the feature subset $\bfS$. Such a property might be desirable for models aiding consequential decision-making as it guarantees good performance under the worst possible target distribution. In other words, the solution to (\ref{eq:fairda_sol}) will perform well for \textit{any} target distribution we may encounter, as long as the distribution adheres to the stated assumptions.

Denote the set of continuous functions which satisfy the fairness constraints $\textbf{G}$ with respect to the distribution $\pd$ by 
$$\mathcal{F}({\textbf{G},\pd}) := \{f\in \mathcal{C}^0 : \textbf{G}(f,\pd)\leq \boldsymbol{\epsilon}\},$$
where $\mathcal{C}^0$ denotes the set of all continuous functions.
Let $\mathcal{P}$ denote the distributions over $(\bfX,A,Y)$ that satisfy Assumptions \ref{par:assum1} and \ref{par:assum2} for some features $\bfS$. Then, the set $\mathcal{F}({\textbf{G},\pd})$ is the same for any distribution $\pd\in\mathcal{P}$.

\begin{lemma}
$\mathcal{F}(G, \pd) = \mathcal{F}(G, Q),\  \forall\ \pd,Q\in \mathcal{P}$
\end{lemma}
By Assumption \ref{par:assum2}, if $\textbf{G}(f, Q)$ holds then $\textbf{G}(f, \pd)$ also holds. Thus, the two sets are the same.
Therefore, we can denote the set of fair functions by $\mathcal{F}(\textbf{G}, \mathcal{P})$. 

For the next result, we will restrict to three fairness definitions (DP, TPR, or TNR) and assume that the conditional outcome, i.e. the random variable $\smash{\pd(Y=1\vert \bfX,A,C=1)}$, has strictly positive density on $[0,1]$. This technical condition allows us to characterize the optimal predictors in $\mathcal{F}(\textbf{G}, \mathcal{P})$, following  \citet{corbett2017algorithmic}.

\begin{theorem}[Worst-case optimality]
\label{thm:advopt}
Consider the set of distributions $\mathcal{P}$ satisfying Assumptions 1 and 2  which are absolutely continuous with respect to the same product measure, and a set of fair functions $\mathcal{F}(\textbf{G}, \mathcal{P})$ satisfying either DP, TPR, or TNR. Assume that the conditional outcome has strictly positive density. Then, the proposed classifier $\tilde{f}^\ast$ satisfies
\begin{equation}
    \tilde{f}^\ast \in \argmin_{f\in \mathcal{F}(\textbf{G}, \mathcal{P})}\  \underset{{\pd\in \mathcal{P}}}{\textup{sup}}\  \pd(f(\bfX, A)\neq Y)
\end{equation}
\end{theorem}
That is, the proposed approach achieves minimum worst-case error amongst the fair predictors with respect to the distributions satisfying the two assumptions. We note that the assumption of absolute continuity in Theorem \ref{thm:advopt} is made to avoid cases where source and target distributions have disjoint support, which would make generalization challenging if some parts of the feature space are not observed at all in the source domain.

\begin{algorithm}[tb]
  \caption{Fair domain adaptation via reduction to standard fair learning}
  \label{alg:fair_da}
\begin{algorithmic}
  \STATE {\bfseries Input} Joint causal graph $\mathcal{G}$, source data $\mathcal{D}_\text{source}$, fairness metric
  \STATE {\bfseries Output} Classifier ${f^*_\text{target}}(\bfS)$ or \texttt{No\_solution}
  \STATE Initialize $R_\text{val}\leftarrow$ \{\}.
  \FOR{$\bfS\subseteq \{\textbf{X},A\}$}
  \STATE Solve $\text{min}_{f \in \mathcal{F}(\bfS)} \pd_\text{source}(f(\bfS)\neq Y)$ and compute error $R_{\text{val}(\bfS)}$ on validation set
  \STATE $R_{\text{val}}\leftarrow \{R_{\text{val}},R_{\text{val}(\bfS)}\}$
  \ENDFOR
  \STATE Sort $R_\text{val}$ in increasing order
  \STATE Traverse $R_\text{val}$ and select $\bfS$ satisfying Assumptions \ref{par:assum1} and \ref{par:assum2}, say $\bfS^*$, by checking for d-separation in graph $\mathcal{G}$
  \IF{$\bfS^*$ exists}
  \STATE Solve Fair DA problem (\ref{eq:fairda_sol}) with features $\bfS^*$ and return output
  \ELSE
  \STATE return \texttt{No\_solution}
  \ENDIF
\end{algorithmic}
\end{algorithm}

\begin{figure*}[t]
\centering
\begin{subfigure}[b]{.3\textwidth}
  \centering
\begin{align*}
            &A \sim \text{Bernoulli}\left(\sigma\left({\color{red}\gamma\cdot\lambda_1\cdot C} + u_1\right)\right)\\
            &R \sim \cn(0,1) + \lambda_2\cdot A + u_2\\
            &Y \sim \text{Bernoulli}\left(\sigma\left(\lambda_3\cdot A + \lambda_4\cdot R + u_3\right)\right)\\
            &T = \lambda_5\cdot Y + \lambda_6\cdot R + \lambda_7\cdot A
            + \cn(0,{\color{red}\gamma\cdot\lambda_8\cdot C}) + u_4\\
            &u_1,u_2,u_3 \sim \cn(0,0.8^2),
            u_4 \sim \cn(0,1.0)\\
            &\smash{(\lambda_1,\lambda_2,\lambda_3,\lambda_4) = (0.2, -0.1, -0.8, 0.8)}\\
            &\smash{(\lambda_5,\lambda_6,\lambda_7,\lambda_8) = (0.8, 0.1, -0.8, 0.2)}\\
            &\gamma \in [0,15]\\
            &\sigma(x)=1/(1+\text{exp}(-x))
        \end{align*}
\caption{Data generating process.}
    \label{fig:synth_dgp}
\end{subfigure}\begin{subfigure}[b]{.33\linewidth}
  \centering
  \includegraphics[scale=0.4]{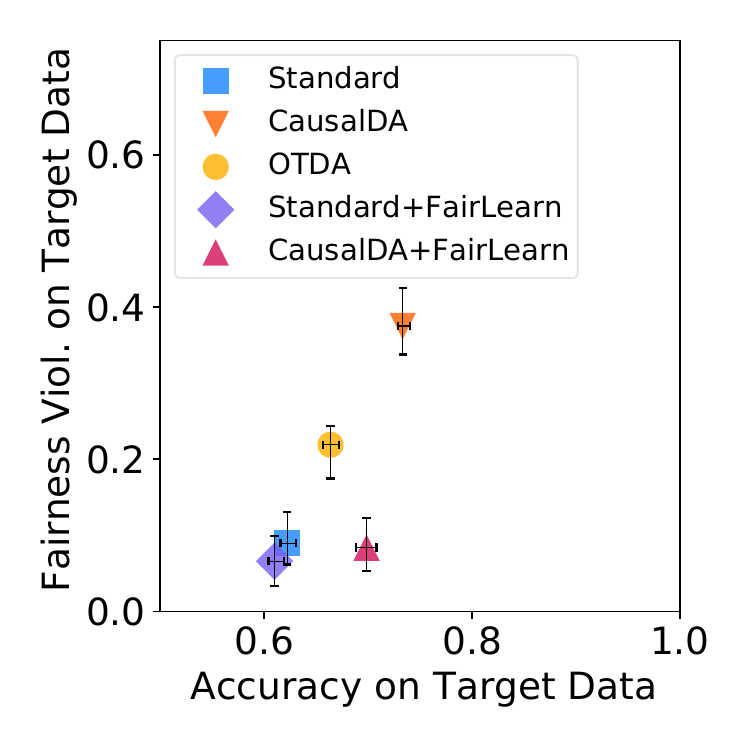}
  \caption{High shift magnitude, $\smash{\gamma=15}$.}
  \label{fig:results_high_diff}
\end{subfigure}\begin{subfigure}[b]{.33\linewidth}
  \centering
  \includegraphics[scale=0.4]{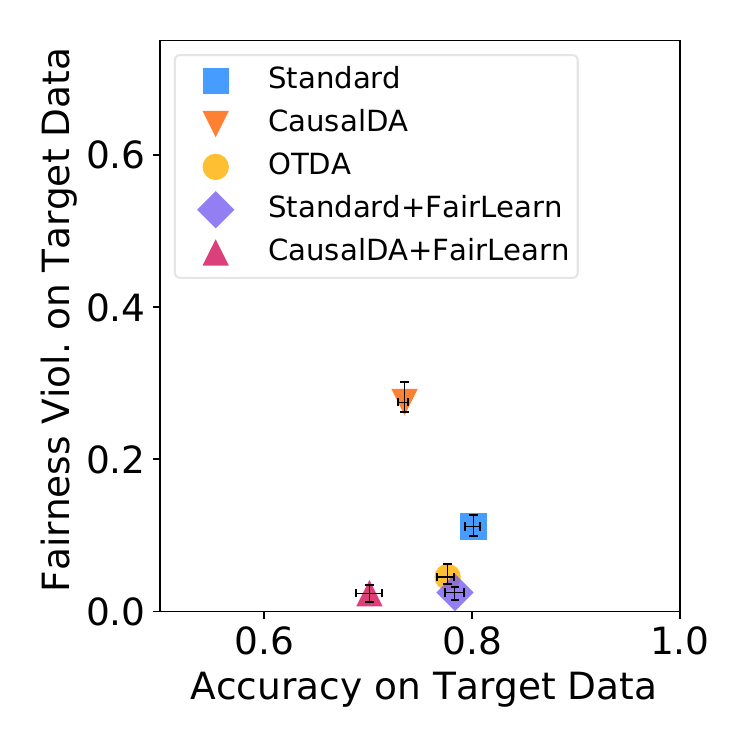}
  \caption{Low shift magnitude, $\smash{\gamma=1.67}$.}
  \label{fig:results_low_diff}
\end{subfigure}
\caption{(a) Data for the domains with shift governed by $\gamma$, highlighted in red. (b,c) Accuracy and fairness metrics on synthetic data example with different magnitude of shifts. Median values are reported over $50$ runs and error bars show first and third quartiles. Proposed approach $\cdaf$ is both accurate and fair under large shifts.}
\label{fig:results_synth}
\end{figure*}

\subsection{Practicality of assumptions}
\label{sec:prac}
Assumptions \ref{par:assum1} and \ref{par:assum2} together describe the types of shifts that our approach can address. Graphically, these are characterized as (a) shifts with causal paths to $Y$ which all include $A$ (i.e. $C {\cdots\rightarrow} A{\rightarrow\cdots} Y$ with all arrows toward $Y$), and (b) shifts with non-causal paths to $Y$ (i.e. $C{\cdots\rightarrow} M{\leftarrow\cdots} Y$ for some feature $M\in\bfS$).
This means that any shift causing change in the distribution of the sensitive attribute as well as any shift in variables with a non-causal path to $Y$ can be addressed. Figure \ref{fig:signature} gives an example of both the cases (described in more detail in Appendix 
\ref{app:addressable}).
Shifts in distribution of sensitive attribute are common when there is sample selection bias e.g. patient demographics being different between rural and urban hospitals.
In Section~\ref{sec:aki}, we demonstrate a general class of shifts in medical diagnosis tasks where both the assumptions are satisfied. Finally, we note that the assumptions (barring those for DP) are untestable without access to labelled target data.  The reason for untestability is the same as that for no unmeasured confounding -- we do not observe the (counterfactual) target data, and hence cannot test for conditional independence. Thus, background knowledge of plausible shifts are critical.

\subsection{Proposed algorithm}

The approach described in (\ref{eq:fairda_sol}) suggests a simple algorithm based on feature selection followed by solving the standard fair learning problem.
We assume that the following are given -- a causal graph for the system of interest $\mathcal{G}$ and data from a source domain $\mathcal{D}_\text{source}=\{(\textbf{X}_i,A_i, Y_i)\}_{i=1}^n$.
The steps, outlined in Algorithm \ref{alg:fair_da}, are as follows.
(a) Iterate over all feature subsets
to rank them in increasing order of their empirical error on the source domain.
(b) Starting from the feature set with the least error, check for Assumptions \ref{par:assum1} and \ref{par:assum2} using $d$-separation \citep{pearl2009causality} in $\mathcal{G}$. 
(c) Solve the fair learning problem with $\mathcal{D}_\text{source}$ limited to model class $\mathcal{F}(\bfS)$. This can be achieved by a fair learning algorithm, such as \citep{agarwal2018reductions}, chosen based on the model class and the fairness definition.
If there is no $\bfS$ satisfying the assumptions, we do not return a solution.

The time complexity is dominated by the search over feature subsets in (a) which is exponential in number of features. To reduce the combinatorial search, we can run a feature selection procedure, e.g. the lasso in case of linear models \citep[Chapter 3]{hastie2005elements}, to prune non-predictive features. Another heuristic is to
start with the set of causal parents of Y (which satisfy Assumption \ref{par:assum1}) and prune it to get a subset satisfying Assumption \ref{par:assum2}.

\subsection{Extension to Counterfactual Fairness}

Another set of fairness definitions based on the causal effect of the sensitive attribute on the prediction have been proposed \citep{kusner2017counterfactual,nabi2018fair,kilbertus2017avoiding}. We consider one version of these \textit{counterfactual} fairness definitions.
\theoremstyle{definition}
\begin{definition}{(Ctf) \citep{kusner2017counterfactual}}
A classifier $\hat{Y}=f(\bfX, A)$ is said to be \textit{counterfactually} fair if the counterfactual distribution of $\hat{Y}$ conditioned on all observed values is the same under $do(A=\texttt{a})$ and $do(A=\texttt{d})$, i.e.
$\pd(\hat{Y}_{do(A=\texttt{a})}=y\vert \bfX=\bfx, A=i)=\pd(\hat{Y}_{do(A=\texttt{d})}=y\vert \bfX=\bfx, A=i)$,
for $y\in\{0,1\}$ and $i\in\{\texttt{a},\texttt{d}\}$.
\end{definition}
One method to build a classifier $f(\bfS)$ satisfying Ctf is to only use feature set $\bfS\in\{\bfX,A\}$ that does not contain any descendant of $A$ in the causal graph \citep[][Lemma 1]{kusner2017counterfactual}.

Thus, the counterpart of Assumption \ref{par:assum2} for solving Fair DA under Ctf is that the selected feature set contains the non-descendants of $A$. Combined with Assumption \ref{par:assum1}, we select non-descendants of $A$ which form a separating set in order to solve Fair DA.
Since, Ctf only requires change in feature subset and does not include any fairness constraints in the fair learning problem, we can show the worst-case optimality result as well (described in Appendix
\ref{app:counterfactual}).

However, we note that there are multiple ways of defining counterfactual fairness. For instance, \citep{nabi2018fair} require that causal effects of $A$ on $\hat{Y}$ through particular paths should be zero or small.
Further work should explore approaches to solve Fair DA under broader definitions of counterfactual fairness. 
\section{Experiments}
\label{sec:exp}

The experiment settings explained next are designed to evaluate performance (accuracy and fairness) of the proposed classifier, trained using a source dataset, on unseen target datasets.
The constrained learning problem in (\ref{eq:fairda_sol}) is solved using the algorithm by \citep{agarwal2018reductions}, referred henceforth as $\fl$, which converts the problem into a sequence of weighted cost-sensitive classification problems.
Predictive performance is measured using accuracy (percentage correct), area under ROC and precision-recall curves (AUPRC). For the experiments presented here, we use EO as the desired fairness constraint. To evaluate (un)fairness, we report the
maximum violation of the EO constraint, i.e.
$\smash{\max_{Y\in\{0,1\},A\in\{\texttt{a},\texttt{d}\}} \big | \pd(f\left(\bfS\right) \mid Y,A) - \pd(f\left(\bfS\right)\mid Y) \big|
}$.

\subsection{Baselines.}
We consider five baselines which account for either distribution shift, unfairness, both, or none of these.
\begin{itemize}
\item $\std$ is the optimal un-constrained classifier with \textit{all} available features, i.e. $\smash{f(\textbf{V}\setminus Y)}$. 
\item $\cda$ is the classifier with the separating set, i.e. $f(\bfS)$ s.t. $\smash{C\independent Y\mid \bfS}$. 
\item $\ot$ is an optimal transport-based method for unsupervised domain adaptation \citep{courty2016optimal}.
\item $\stdf$ is $f(\textbf{V}\setminus Y)$ trained with $\fl$. 
\item Finally, $\cdaf$ is the proposed method i.e. $f(\bfS)$ trained with $\fl$ where $\bfS$ satisfies Assumptions \ref{par:assum1} and \ref{par:assum2}.
\end{itemize}
Results on another method, \textit{anchor regression} \citep{rothenhausler2018anchor}, are included in Appendix 
\ref{app:addexp}. Since this method requires data from multiple sources, we evaluate it against the above methods in a separate experiment setting.
Hyperparameters used for the methods are reported in Appendix 
\ref{app:hyperparam}. Code for reproducing results on synthetic data is at \url{https://github.com/ChunaraLab/fair_domain_adaptation}.

\begin{figure*}[t]
\centering
\begin{subfigure}[b]{.33\linewidth}
\centering
        \scalebox{0.9}{\begin{tikzpicture}

    \node[state,rectangle,fill=blue!10] (c) at (0,0) {$C_1$}; \node[state] (d) [below =of c] {$D$}; \node[state] (y) [below left =of d] {$Y$}; \node[state] (x) [below right =of y] {$X$}; 
    \node[state] (c_1) [below right =of d]{$M$}; \node[state,rectangle,fill=blue!10] (c_2) [right =of x] {$C_2$};

 \path (c) edge (d);
    \path (d) edge (y);
    \path (d) edge (x);
    \path (d) edge (c_1);
    \path[bidirected]  (x) edge[bend left=30] (y);
\path (c_1)  edge (x);
    \path (c_1) edge (y);t
    \path (c_2) edge (x);

\end{tikzpicture} }
        \caption{Graph for AKI diagnosis.}
        \label{fig:real}
\end{subfigure}\begin{subfigure}[b]{.33\linewidth}
\centering
    \begin{tabular}{ |c|c| }
     \hline
     D & \specialcell{Demography\\(age, sex (A), race)} \\
     \hline
     Y & AKI diagnosis \\ 
     \hline
     M & Comorbidities\\ 
     \hline
     X & \specialcell{Lab tests \& Vitals\\(including BUN)}\\ 
     \hline
     $C_1, C_2$ & Context variables\\
     \hline
    \end{tabular}
    \caption{Variable descriptions}
\label{fig:real_variables}
\end{subfigure}\begin{subfigure}[b]{.33\linewidth}
  \centering
  \includegraphics[scale=0.33]{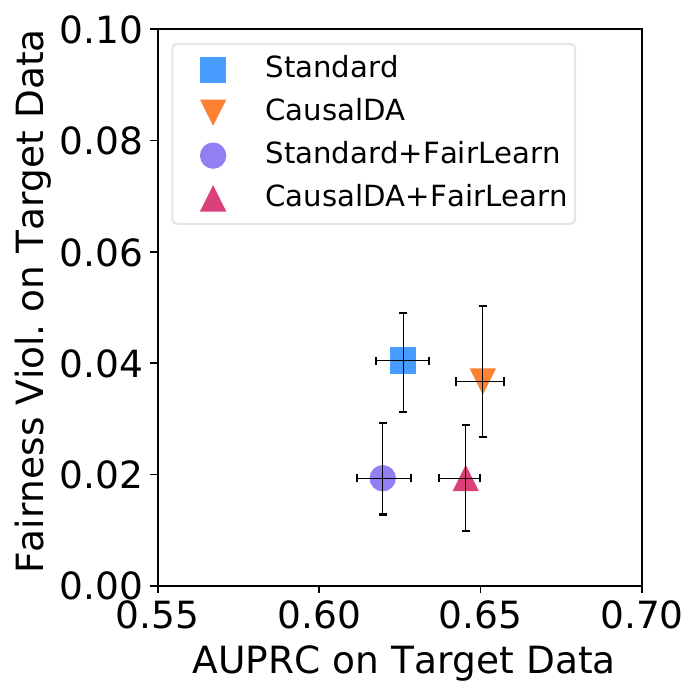}
  \caption{AUPRC. Class ratio is $0.21$}
  \label{fig:results_aki_auprc}
\end{subfigure}
\caption{(a) Postulated causal graph for AKI. Bi-directed edge denotes unmeasured confounding between disease outcome and lab test values due to unobserved common causes.
(b) Legend for variables in the graph.
(c) Accuracy and fairness metrics for AKI data. Median values are reported over $50$ runs and error bars show first and third quartiles. Proposed approach improves fairness with small loss in accuracy, even on shifted target data.}
\label{fig:results_aki}
\end{figure*}
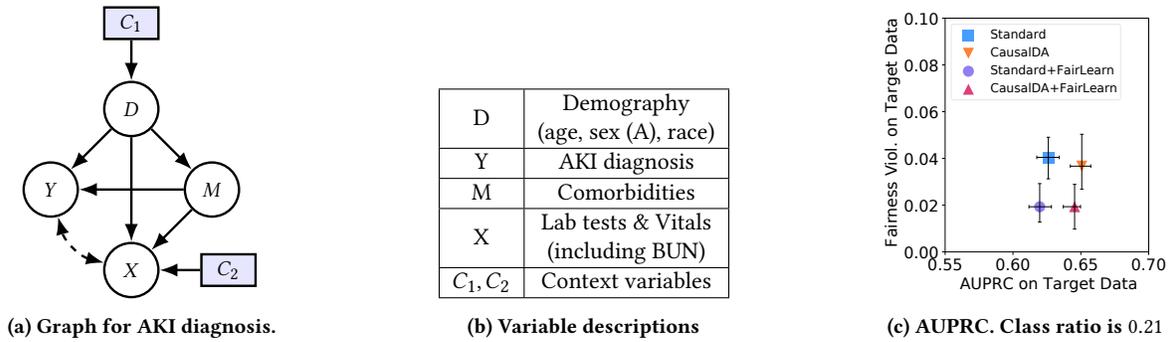

\subsection{Synthetic data example}
\label{sec:synth_exp}

\noindent
\textbf{Setup.}
For the flu example in Figure~\ref{fig:synth_data_graph}, we generate data from a structural equation model described in Figure \ref{fig:synth_dgp} with linear relationships and logit link function for binary variables. To generate target domains, we perform soft interventions \citep{markowetz2005probabilistic} to shift distributions of $A$ and $T$. The shift magnitude is governed by a multiplier $\gamma$ in the linear equations.
In total, $50$ pairs of source and target datasets are simulated with $N=2000$ samples in each dataset. The proportion of disadvantaged group in source is kept at roughly $0.5$. In target domains with an extreme value for $\gamma=15$, the ratio shifts to roughly $0.94$. Class ratio is varied from $0.5$ to $0.36$ with increase in $\gamma$.
From Figure~\ref{fig:synth_data_graph}, we observe that $\bfS{=}\{A,R\}$ satisfies the two assumptions. 
Adding $T$ (a collider) to $\bfS$ makes the predictor dependent on $C$ and, thus, unstable. 
We use logistic regression models in all experiments.
In Figure \ref{fig:results_high_diff}, the goal is to find a classifier performing well on both accuracy and fairness, i.e. one close to the right-hand bottom corner.

\textbf{Results.} For a high magnitude of shift, Figure \ref{fig:results_high_diff}, domain adaptation ($\cda$) leads to considerably higher accuracy than using all features ($\std$), but results in high unfairness.
Learning with fairness constraints ($\cdaf$) which results in low unfairness with a minimal loss in accuracy even when the domains differ significantly.
As seen in Figure \ref{fig:results_low_diff}, for low magnitudes of shift, $\cdaf$ still has low unfairness but results in a pessimistic accuracy estimate as it accounts for larger shifts than are seen in the target domain. 
Thus, in practice, the choice of method will depend on the expected magnitude of shift.

\subsection{Synthetic data example: additional results}

\begin{figure*}[t]
\centering
\begin{subfigure}{.33\linewidth}
  \centering
  \includegraphics[scale=0.33]{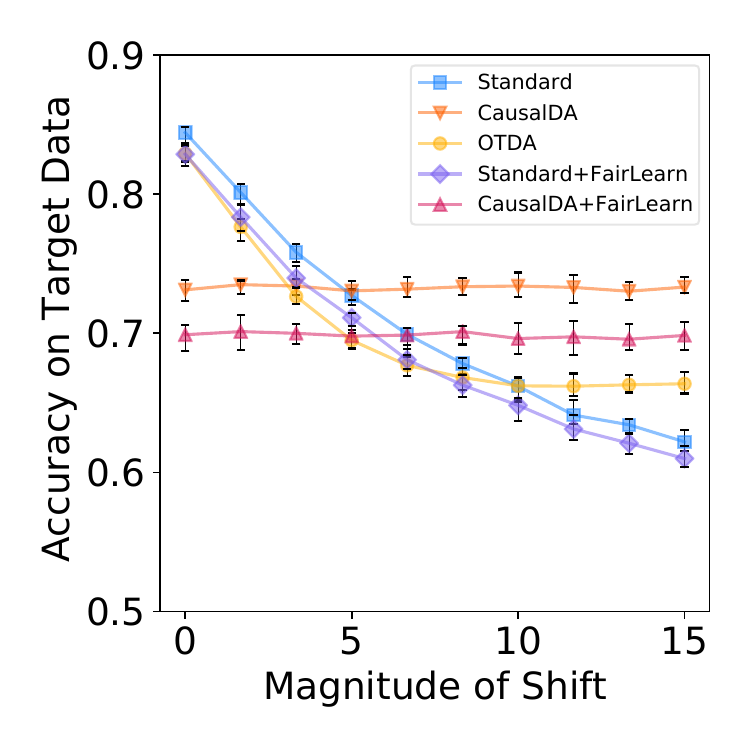}
  \caption{Accuracy vs Shift.}
  \label{fig:results_synth_acc_all}
\end{subfigure}\begin{subfigure}{.33\linewidth}
  \centering
  \includegraphics[scale=0.33]{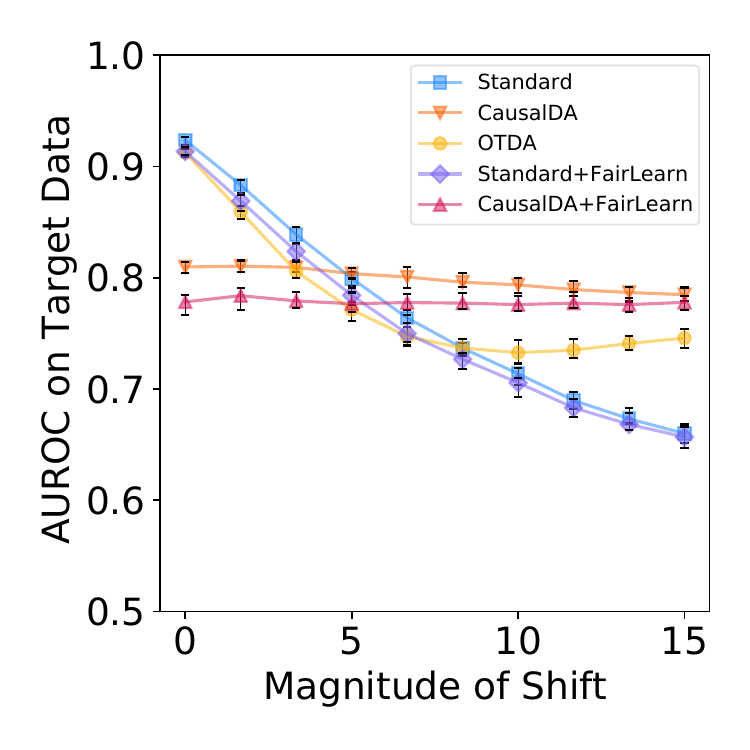}
  \caption{AUROC vs Shift.}
  \label{fig:results_synth_roc_auc_all}
\end{subfigure}\begin{subfigure}{.33\linewidth}
  \centering
  \includegraphics[scale=0.33]{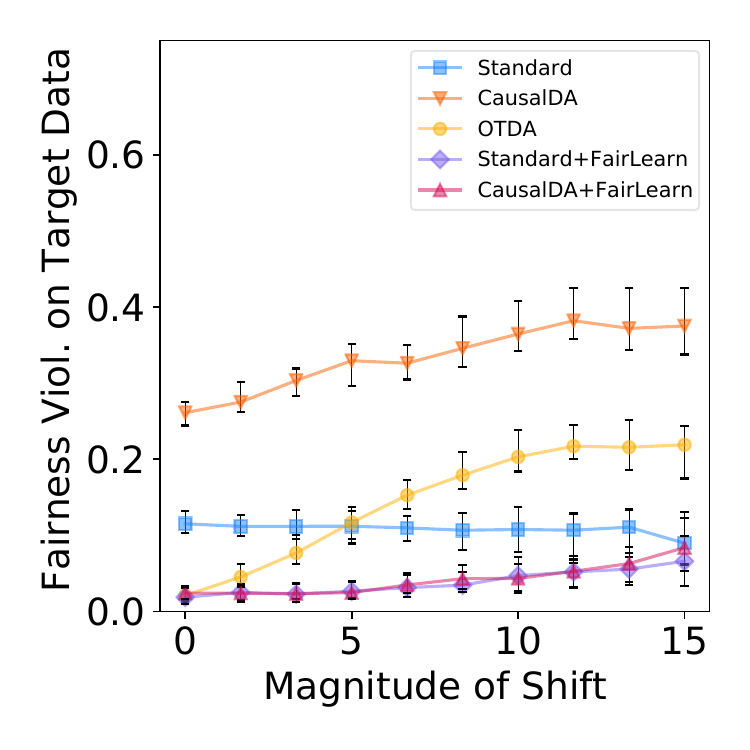}
  \caption{Fairness Violation vs Shift.}
  \label{fig:results_synth_fair_all}
\end{subfigure}
\begin{subfigure}{.33\linewidth}
  \centering
  \includegraphics[scale=0.33]{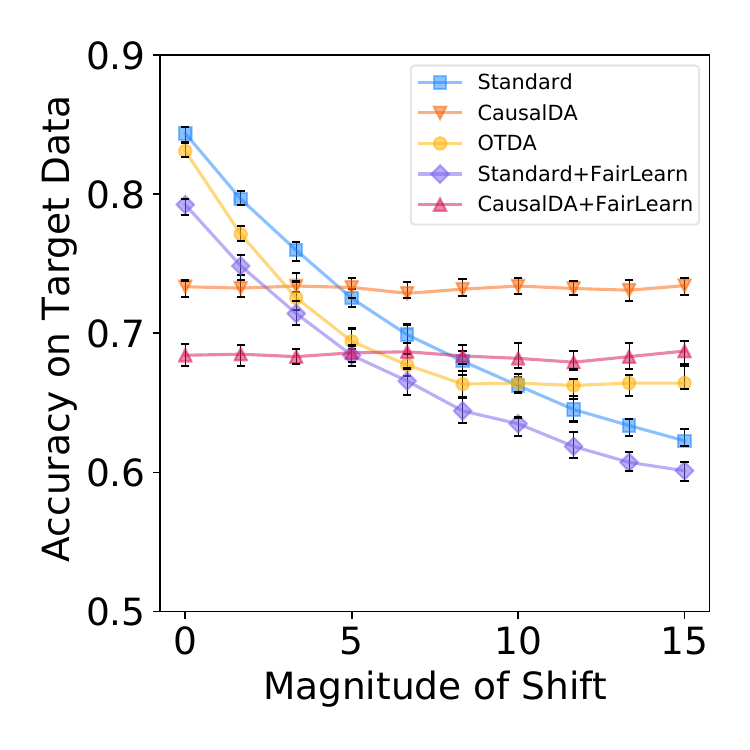}
  \caption{Accuracy vs Shift.}
  \label{fig:results_synth_acc_dp_all}
\end{subfigure}\begin{subfigure}{.33\linewidth}
  \centering
  \includegraphics[scale=0.33]{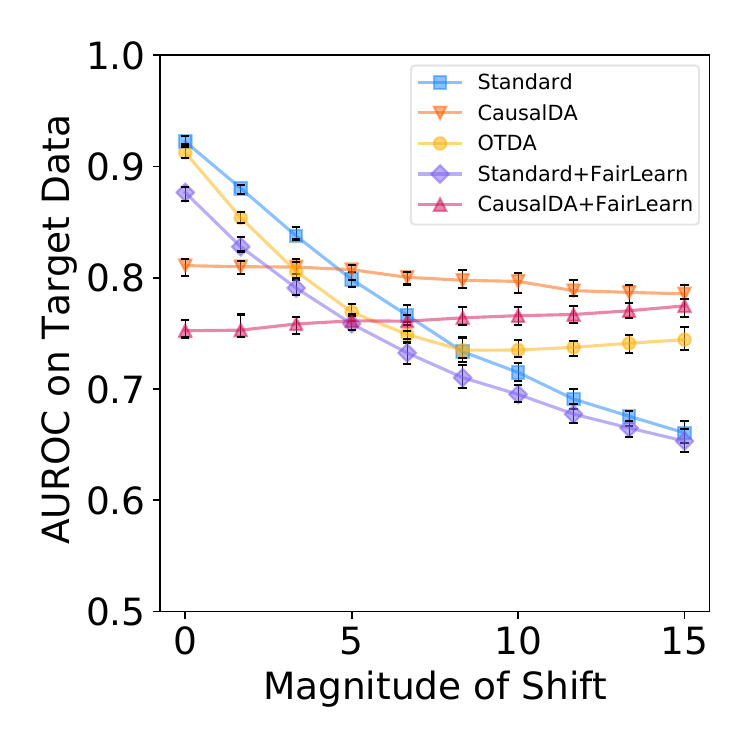}
  \caption{AUROC vs Shift.}
  \label{fig:results_synth_roc_auc_dp_all}
\end{subfigure}\begin{subfigure}{.33\linewidth}
  \centering
  \includegraphics[scale=0.33]{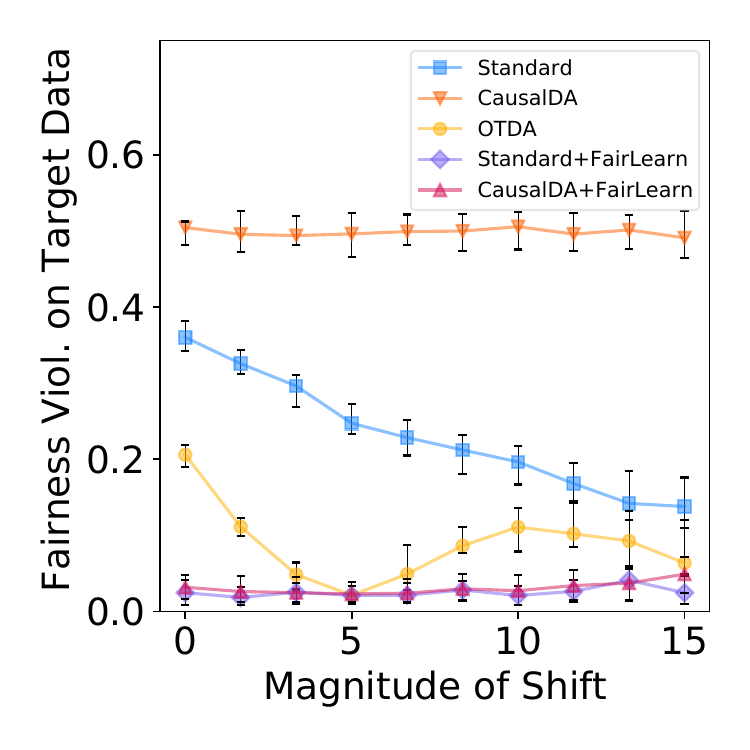}
  \caption{Fairness Violation vs Shift.}
  \label{fig:results_synth_fair_dp_all}
\end{subfigure}
\caption{Accuracy, AUROC, and Fairness violation with varying magnitude of shifts for synthetic data. (a,b,c) With equalized odds (EO) as the fairness constraint. (d,e,f) With demographic parity (DP) as the fairness constraint. Median values are reported over $50$ runs and error bars show first and third quartiles. Performance of the proposed approach is stable across different shifts and for the two fairness metrics.}
\label{fig:results_synth_all}
\end{figure*}

\noindent
\textbf{Varying magnitude of shift.}
To check robustness of different models to distribution shift, we generate target datasets with different values of $\gamma$ in the linear structural equations in Section 6.1. Figure \ref{fig:results_synth_all} (a,b,c), included at the end, shows two predictive performance metrics -- Accuracy (percentage correct), AUROC -- and one fairness metric -- maximum fairness violation -- for different magnitudes of shift. We observe the same trends as reported in Section 6.1, i.e. $\cda$ (orange curve) achieves stable predictive performance but leads to high unfairness, whereas $\cdaf$ (red curve) achieves both stable predictive performance and low unfairness.

\textbf{Results with demographic parity.}
Figure \ref{fig:results_synth_all} (d,e,f) report results on the synthetic example with demographic parity (DP) as the fairness constraint instead of EO. In case of DP, the fairness violation is quantified as
$\smash{| \pd(f\left(\bfS\right) \mid A=a) - \pd(f\left(\bfS\right)\mid A=d) \big|
}$.
We observe similar trends as compared to the plots for EO.

\subsection{Case study: diagnosing Acute Kidney Injury}
\label{sec:aki}

Acute Kidney Injury (AKI) is a condition characterized by an acute decline in renal function, affecting 7-18\% of hospitalized patients and more than $50$\% of patients in the intensive care unit (ICU)~\citep{chawla2017acute}. 
The condition can develop over a few hours to days and early prediction can greatly reduce the fatalities associated with the condition. Hence, building models for predicting AKI risk from clinical data is an active area of research. Such models can be used to risk-stratify patients to screen them for close monitoring or to perform further diagnostics to guide course of treatment \citep{hodgson2019role}.
Importantly, AKI incidence has well-documented disparities across groups defined by race and sex \citep{grams2014explaining, grams2015meta}. Thus, introduction of risk prediction tools for guiding clinical care has a potential to perpetuate such disparities, or alternatively, to address them through a more deliberate design of the prediction tools.
A recent study \citep{tomavsev2019clinically} showed good predictive performance for AKI based on patient data provided by the U.S. Department of Veteran Affairs. However, the female population was severely underrepresented in the data, which raises concern over differential error rates when deployed in a different population. 
Therefore, to analyze the fairness across sensitive groups, we conduct experiments on MIMIC III, a publicly-available critical care dataset \cite{johnson2016mimic}.
We extract variable types, mentioned in caption of Figure~\ref{fig:results_synth}, for around $24K$ patients. Pre-processing steps are described in Appendix
\ref{app:akipreprocess}.
We use a simplified causal graph for the AKI diagnosis task, Figure \ref{fig:real}, based on the one used by \cite{subbaswamy2018counterfactual} for a sepsis diagnosis task. The group \textit{sex=female} is taken as the sensitive attribute to assess fairness of the predictions. 
In this case study, the AKI risk score is not intended to prescribe treatment, but to flag a patient for extra care resources e.g. by alerting clinical staff. Thus, the potential harm that we want to avoid is groups having unequal opportunity to such care resulting from group differences in prediction errors.

\textbf{Setup.} Patient encounters are randomly split (2:1) into source and target data.
We artificially introduce two types of shifts
-- (a) change in female proportion, and (b) change in measurement policy, where a lab test is prescribed less often -- some of the factors affecting model performance across clinical settings \citep{riley2016external}.
We randomly downsample female population by rejecting each row in the source data from that group with probability $50\%$. This shifts the proportion of females from $40\%$ to $25\%$.
Also, we randomly choose $50\%$ encounters in the target data and add missing values for the Blood Urea Nitrogen (BUN) test, a biomarker of AKI \citep{edelstein2008biomarkers}.
Results with other missingness proportions are included in Appendix
\ref{app:propmissing}.

From Figure \ref{fig:real}, we note that $\bfS{=}\{\textbf{D},\textbf{M},\bfX\setminus \text{BUN}\}$ satisfies the two assumptions. 
We report AUPRC in Figure~\ref{fig:results_aki_auprc}, instead of accuracy, as it is less sensitive to class imbalance (class ratio is $0.21$). 
All results are reported for classifiers trained with gradient boosting trees. We drop $\ot$ from comparison due to its low accuracy and high running time for this dataset.

\textbf{Results.} We find that classifiers with separating feature set perform significantly better in AUPRC compared to those with all features (exact numbers are reported in Appendix
\ref{app:propmissing}). 
Further, $\cdaf$ improves fairness in target domain, reducing fairness violation by $47\%$ with $0.8\%$ decrease in AUPRC.
Thus, the experiments provide preliminary evidence that our method can learn stable classifiers while being fair for a class of shifts in diagnosis tasks denoted by Figure~\ref{fig:real}. Note that the setup has some limitations, namely, adding missing values to perturb target data conflates the effectiveness of the procedure for handling missing data (mean value imputation in our case) with the procedure for domain adaptation. 
We plan to validate the approach on datasets across multiple hospitals or time points to address these limitations.

\section{Limitations and Discussion}

\noindent
\textbf{Knowledge of causal graph.}
Our approach requires the causal graph for the system being studied to check whether the two assumptions are satisfied for any given subset. While this is a requirement made by multiple domain adaptation methods \citep{subbaswamy2019preventing, subbaswamy2018counterfactual}, this can be relaxed when data from multiple domains are available. In such settings, causal discovery methods \citep{peters2016causal} can be used to posit a graph and validate with domain experts. Such a procedure is demonstrated in \citep{subbaswamy2020spec}.
An important direction for future work includes identifying the desired feature subsets with causal discovery algorithms.
We recommend that the causal graph be postulated conservatively, i.e. only adding conditional independencies that are well-substantiated by domain knowledge. In this case, if the separating features are not found, our method will output that a fair predictor is not possible instead of incorrectly returning a model that will not be fair.

\textbf{Addressable shifts.}
In Section \ref{sec:prac}, we described shifts that our approach can address and presented examples.
However, these are only a part of the possible shifts that a modeller may worry about. For example, shifts in direct causes of the outcome are excluded due to Assumption \ref{par:assum2}. Such shifts can result in arbitrary changes to the outcome within each group, making it impossible to balance group-specific statistics in the fairness constraint (see Appendix 
\ref{app:example} for an example). These are difficult to address without making strict assumptions on magnitude of the shift or assuming access to target data. Thus, for some joint causal graphs, Algorithm \ref{alg:fair_da} might not yield any feature set. 
In such cases, an alternative is to return the set with the least source domain risk but such a set has no generalization guarantee.

\textbf{Algorithmic fairness in healthcare to promote health equity.} Disparities in health outcomes and healthcare access across different groups (e.g. based on race and gender) arise from multiple reasons such as socio-economic inequities (e.g. due to structural racism) \citep{phelan2015racism} and clinician bias \citep{iom2003unequal}. Such disparities can result in differential model performance across groups as \citet{obermeyer2019dissecting} finds in context of a model for identifying patients who need extra care resources. Left unaddressed, allocating resources using `biased' models may worsen health disparities. As a consequence, a growing body of work aims to develop algorithms embodying fairness principles specific to healthcare \citep{chen2020ethical}. This includes constraining prediction errors across groups for the tasks of predicting risk of cardiovascular events \citep{pfohl2019cardiovascular} or predicting healthcare costs \citep{zink2020fair}.
However, such group-level fairness constraints, including the ones we consider, may not match ethical desiderata in all possible healthcare settings. Some alternative constraints have been defined, for example, using counterfactuals \citep{pfohl2019counterfactual} or preference between group or aggregate-level models \citep{ustun19preference}. We plan to investigate fair domain adaptation under broader notions of fairness. 
We have motivated the approach on healthcare tasks due to the importance of ensuring reliable model performance under distribution shifts in this domain. We note that the approach is more broadly applicable to other domains involving high-stakes decisions.

\section{Conclusion and future work}
In absence of data from new environments in which a machine learning model will be deployed, giving performance guarantees regarding predictive performance and fairness is challenging. 
We find that methods to address distribution shift, while controlling for decay in accuracy, can result in fairness violations.
As a counter-measure, we show that it is possible to obtain accurate and fair predictors for widely-studied fairness definitions and under a large class of shifts particularly prevalent in healthcare tasks.
Future work includes studying fair domain adaptation under parametric assumptions on shifts, adaptation for counterfactual definitions of fairness, and finite sample properties of the estimators.
We hope that the problem setup presented here will enable further work at the intersection of fairness and causal inference. 
\begin{acks}
We acknowledge funding from the NSF grant number 1845487. HS would like to thank Sreyas Mohan, Kunal Relia, Margarita Boyarskaya and Nabeel Abdur Rehman for helpful discussions.
\end{acks}

\bibliographystyle{ACM-Reference-Format}
\bibliography{transfer_learning}

\clearpage

\appendix

\section{Proposition 1: Example with group-specific measurement error}
\label{app:example}

\begin{proof}[Proof of Proposition 1]
Through a simple example, we show that fair domain adaptation (DA) is not possible without making assumptions, even in cases where domain adaptation is possible.

Consider the data generating process, represented by Figure \ref{fig:non_conforming}, with two sensitive groups $A\in\{0,1\}$, a covariate $X$, and an outcome $Y\in\reals$.

\begin{figure}[htbp!]
\centering
    \begin{subfigure}[b]{0.5\columnwidth}
        \centering
        \begin{align*}
            A &\sim \text{Bernoulli}(0.5),\\
            X &= A + \gamma C \times A + \epsilon_X,\\
            Y &= X + A + \epsilon_Y,\\
            \epsilon_X,\epsilon_Y &\sim \cn(0,1).
        \end{align*}
        \caption{Data generating process}
    \end{subfigure}\begin{subfigure}[b]{0.5\columnwidth}
        \centering
        \begin{tikzpicture}

    \node[state,rectangle,fill=blue!10] (c) at (0,0) {$C$}; \node[state] (x) [below =of c] {$X$}; \node[state] (y) [right =of x] {$Y$}; \node[state] (d) [ right =of c] {$A$};

\path (c) edge (x);
    \path (x) edge (y);
\path (d) edge (x);
    \path (d) edge (y);

\end{tikzpicture}          \caption{Causal graph}
    \end{subfigure}
\caption{Example with group-specific measurement error. The group-specific errors, governed by $\smash{\pd(X\mid A,C)}$ in case of DP, can change arbitrarily between the domains depending on the mechanism for $C\rightarrow X$. Thus, we cannot constrain the difference in group-specific errors to satisfy DP in target. For this example, the proposed method fails to find a feature set satisfying Assumptions 1 and 2.}
\label{fig:non_conforming}
\end{figure}
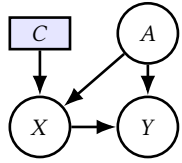

That is, the feature $X$ for the subgroup $A=1$ is corrupted in the target domain $C=1$. The magnitude of corruption is governed by a constant $\gamma$ that depends on the particular target domain of interest.

Suppose, we want to build a predictor, $\hat{Y}=f(X,A)$, satisfying demographic parity (DP) as the fairness definition. DP requires $f(X,A)\independent A$. Thus, the fairness constraint is $G^{C=1}(f)=\vert \ep(f(X,A)\mid A=1, C=1)-\ep(f(X,A)\mid A=0, C=1)\vert$. Define the mean squared error in target domain as $L^{C=1}(f)=\ep_{\pd(X,A,Y\vert C=1)}((f(X,A) - Y)^2)$.

Then, the fair DA problem requires finding a predictor for the target domain $C=1$, given data from the source domain $C=0$, s.t.

\begin{equation}
\label{eq:fair_da_prop1}
    f^\ast=\argmin_{f \in \mathcal{F}} \{L^{C=1}(f) : G^{C=1}(f)\leq \epsilon\}.
\end{equation}

We will restrict to linear models as the true distribution lies in the class of linear models. 
Let, $f(X,A;\beta)=\beta_0 + \beta_1 X + \beta_2 A$, for some unknown vector $\beta$.

Then, the fairness constraint is given by $G^{C=1}(f)=\vert\beta_1 + \beta_1 \gamma + \beta_2\vert$.\\
Note that $G^{C=1}(f)$ depends on the domain $C=1$ through $\gamma$.
That is, the fairness constraint $G^{C=1}(f)$ can change arbitrarily depending on the value of $\gamma$ for the particular target domain, while the source data distribution $C=0$ available to us is fixed. In other words, the fairness constraint (thought of a function of the target distribution $\pd(X,A,Y\vert C=1)$) is not identifiable by the observed distribution $\pd(X,A,Y\vert C=0)$ alone. Thus, we cannot solve (\ref{eq:fair_da_prop1}) without further assumptions or without data from the target domain.
\end{proof}

We constructed the example with a regression task but the same can be shown for a classification task e.g. a logistic function for $f(X,A)$. The DP constraint will again depend on $\gamma$, in general.

The example also motivates the Assumption 2 for DP which requires identifying a feature set $\bfS$ s.t. $C\independent \bfS\mid A$ as it guarantees $G^{C=1}(f)=G^{C=0}(f)$. With this assumption along with the separating set assumption, we can find favourable solutions for Fair DA as shown in Section 5. 
\section{Proposition 2: Solution using data re-weighting}
\label{app:reweighting}

\begin{proof}[Proof of Proposition 2]
To show this, we consider the terms involved in Fair DA (\ref{eq:fair_da_context}), namely classification error and fairness constraint, separately. 
\begin{align}
\label{eq:fair_da_context}
    \argmin_{f \in \mathcal{F}(\bfS)}\  \{\pd(f(\bfS)\neq Y\vert C=1) : \textbf{G}(f, \pd(Y,\bfS\vert C=1))\leq \boldsymbol{\epsilon}\}
\end{align}

We will show that the two terms can be computed by labelled source data, sampled from $P(\bfS,A,Y\vert C=0)$, and unlabelled target data, sampled from $P(\bfS,A\vert C=1)$. Thus, showing that (\ref{eq:fair_da_context}) can be indirectly solved.

The classification error can be written as,
\begin{align}
    \pd(f(\bfS)\neq Y\vert C=1) &= \ep_{\pr(\textbf{V}\vert C=1)}(\ind(f(\bfS)\neq Y))\nonumber\\
    &= \ep_{\pr(Y,\bfS\vert C=1)}(\ind(f(\bfS)\neq Y))\label{eq:overall_risk_1}\\
    &= \ep_{\pr(\bfS\vert C=1)} \left(\ep_{\pr(Y\vert \bfS, C=1)} (\ind(f(\bfS)\neq Y))\right)\label{eq:overall_risk_2}\\
    &= \ep_{\pr(\bfS\vert C=1)} \left(\ep_{\pr(Y\vert \bfS, C=0)} (\ind(f(\bfS)\neq Y))\right)\label{eq:overall_risk_3}\\
    &=\ep_{\pr(\bfS\vert C=1)} \left(\pd(f(\bfS)\neq Y\vert \bfS, C=0)\right)\nonumber\\
    &=\ep_{\pr(\bfS\vert C=0)} \left(\frac{\pr(\bfS\vert C=1)}{\pr(\bfS\vert C=0)}\pd(f(\bfS)\neq Y\vert \bfS, C=0)\right)\label{eq:overall_risk_4}
\end{align}
Here, (\ref{eq:overall_risk_1}) marginalizes out features $\textbf{V}{\setminus} \{\bfS,Y\}$ from the target data distribution as they do not change the error. Step (\ref{eq:overall_risk_2}) follows from the law of iterated expectations, (\ref{eq:overall_risk_3}) uses the conditional independence for the separating set (Assumption 1), and (\ref{eq:overall_risk_4}) uses the importance-weighting identity. We observe that the expectation in \eqref{eq:overall_risk_4} can be estimated just from the available data consisting of $Y$ in the source and $\bfS$ in both the domains. We can re-weight the per-sample source error by the density ratio, $\pr(\bfS\vert C{=}1)/\pr(\bfS\vert C{=}0)$, and take the sample average. The density ratio can be computed, for example, using a probabilistic classifier to discriminate between the domains \citep{bickel2007discriminative}.

Next, consider the fairness constraint. We show the following for EO and the corresponding Assumption 2. The argument is similar for the other definitions i.e. DP, TPR, and TNR. 

Assumption 2 for EO, i.e. $C\independent \bfS\vert Y, A$, implies that $C\independent f(\bfS)\vert Y, A$, assuming that classifier $f$ is a measurable function (which is the case for most classification functions and feature spaces used in machine learning). 
Thus, for $y\in\{0,1\}$, we can write the EO constraints as,
\begin{align*}
    &G(f, \pd(Y,\bfS\vert C=1))\\
    &= \vert\pd(f(\bfS)\neq y\vert Y=y, A=a, C=1)-\pd(f(\bfS)\neq y\vert Y=y, A=d, C=1)\vert\\
    &= \vert\pd(f(\bfS)\neq y\vert Y=y, A=a, C=0)-\pd(f(\bfS)\neq y\vert Y=y, A=d, C=0)\vert\\
    &= G(f, \pd(Y,\bfS\vert C=0))
\end{align*}
Therefore, Assumption 2 guarantees that the evaluation of the fairness constraint does not change with the domain.
The same is true for the other three fairness constraints. For DP, the conditioning on $Y$ is dropped in Assumption 2. Similarly, TPR and TNR fix $Y=1$ and $Y=0$ respectively.

Since both the error and the constraint are estimable, we can minimize (\ref{eq:fair_da_context}) without access to labelled target domain data, if we can find a feature subset satisfying the two assumptions.
\end{proof}

\section{Theorem 5.2: Worst-case optimality}
\label{app:worstcase}

Now, we show that the fair classifier learned in our setting is worst-case optimal. We will restrict to one of the following fairness constraints -- demographic parity (DP), true positive rate equality (i.e. equality of opportunity or TPR), or true negative rate equality (TNR). This restriction enables us to characterize the Bayes optimal classifier under fairness constraints based on results from \citep{corbett2017algorithmic}.

For brevity, we change the notation to represent sensitive attribute $A$ as one of the features in $\bfX$, i.e. system variables are $\textbf{V}:=\{\bfX,Y\}$. In addition, lowercase letters denote observations of random variables which are denoted by the corresponding uppercase letters. For example, $a$ is used to represent the observed value of $A$ (which can be either disadvantaged or advantaged group).

\subsection{Existing result on optimal fair classifier}
The optimal classifier minimizing classification error under either of the fairness constraints (DP, TPR, or TNR) is a threshold function on the conditional outcome probability. In \citep{corbett2017algorithmic}, authors use the term statistical parity for DP and predictive equality for false positive rate equality (or false negative rate equality) which is the same as ensuring TNR (or TPR). Formally, they prove the following.

\begin{lemma}[Theorem 3.2 by \citet{corbett2017algorithmic}]
\label{lem:trule}
Suppose we want to find a classifier $f$ that maximizes \textit{immediate utility}, defined as $U(f,c)=\ep(yf(\bfx)-cf(\bfx))$ for a constant $c\in (0,1)$, and satisfies the fairness constraint (either DP, TPR, or TNR). Let, $p_{y\vert \bfx}(\bfx)=\pr(Y=1\vert \bfX=\bfx)$. Assume that the distribution of the random variable $p_{y\vert \bfx}(\bfX)$ has positive density on $[0,1]$. Then, the optimal classifier $f^\ast$ under the fairness constraint is a threshold function i.e. $f^\ast(\bfx)=\ind[p_{y\vert \bfx}(\bfx)\geq t_a]$ for some constant thresholds $t_a\in [0,1]$ that can optionally depend on the sensitive attribute $A=a$.
\end{lemma}
The result also holds when the fairness constraints are required to be only approximately satisfied.

To relate this lemma to our problem, note that our objective of minimizing classification error is same as maximizing immediate utility for $c=0.5$, as observed by \citet[][Lemma 1]{lipton2018does}. To see this, rewrite the error as $\pd(f(\bfx)\neq y) = \ep(\ind(f(\bfx)\neq y)) = \ep(1-y f(\bfx)-(1-y)(1-f(\bfx)))$ for binary labels $y\in\{0,1\}$.
Rearranging terms, we get 
\begin{align*}
    \pd(f(\bfx)\neq y) &= \ep(1-yf(\bfx)-(1-y)(1-f(\bfx)))\\
    &= \ep(-2yf(\bfx)+f(\bfx))+\ep(y)\\
    &= -2U(f,0.5)+\ep(y)
\end{align*}
where $\ep(y)$ is a constant. Thus, we can use Lemma \ref{lem:trule} to characterize the optimal classifier under classification error and the three fairness definitions.

\subsection{New result on optimal fair classifier under distribution shift}
Now we describe the setting in which the new result is established.

Let $\mathcal{P}$ be the set of distributions over $(\bfX, Y)$ that satisfy Assumptions 1 and 2. That is, for all distributions in $\mathcal{P}$, there exists a feature subset $\bfS$ s.t. $P(Y\vert \bfS)$ and $P(\bfS\vert A)$ (for DP, as an example) are invariant. Let the source distribution be denoted by $\mathbb{Q}\in \mathcal{P}$. In addition, assume that the distributions in $\mathcal{P}$ are absolutely continuous with respect to the same product measure.

The proposed classifier is $\tilde{f}^\ast$ that satisfies
\begin{align*}
    \tilde{f}^\ast \in & \argmin_f E_\mathbb{Q}(\ind[y\neq f(\bfs)]),\\
    & \text{s.t. } \textbf{G}(f, \mathbb{Q}) \leq \boldsymbol{\epsilon},
\end{align*}

Denote the set of continuous functions which satisfy the fairness constraint $\textbf{G}$ w.r.t. the source distribution $\mathbb{Q}$ by $\mathcal{F}(\textbf{G}, \mathbb{Q}) := \{f\in \mathcal{C}^0 \vert \textbf{G}(f,\mathbb{Q}) \text{ holds}\}$. Then, we show that the set is the same for all distributions in $\mathcal{P}$.

\begin{lemma}
$\mathcal{F}(\textbf{G}, \mathbb{Q}) = \mathcal{F}(\textbf{G}, \mathbb{P}),\  \forall\ \mathbb{P}\in \mathcal{P}$
\end{lemma}
\begin{proof}
Under Assumption 2, the fairness constraints are invariant. Thus, if $\textbf{G}(f, \mathbb{Q})$ holds then $\textbf{G}(f, \mathbb{P})$ also holds for any distribution $\mathbb{P}\in \mathcal{P}$ and \textit{vice versa}.
\end{proof}

Thus, we will denote the set of fair functions by $\mathcal{F}(\textbf{G}, \mathcal{P})$. The proposed estimator can be written as
\begin{align*}
    \tilde{f}^\ast \in & \argmin_{f\in \mathcal{F}(\textbf{G}, \mathcal{P})} E_\mathbb{Q}(\ind[y\neq f(\bfs)])
\end{align*}

We will show that the proposed predictor $\tilde{f}^\ast$ is optimal over the set of fair functions w.r.t. plausible target distributions in $\mathcal{P}$ and fairness constraint $\textbf{G}$ in an adversarial setting.

\begin{theorem}
\label{thm:worst_case_optimal}
Consider the set of distributions $\mathcal{P}$ satisfying Assumptions 1 and 2 which are absolutely continuous with respect to the same product measure, and a set of fair functions $\mathcal{F}(\textbf{G}, \mathcal{P})$ satisfying either DP, TPR, or TNR. Assume that the conditional outcome has strictly positive density. Then, the proposed classifier $\tilde{f}^\ast$ satisfies
\begin{equation}
\label{eq:worst_case}
    \tilde{f}^\ast \in \argmin_{f\in \mathcal{F}(\textbf{G}, \mathcal{P})}\  \underset{{\mathbb{P}\in \mathcal{P}}}{\text{sup}}\  \ep_\mathbb{P}(\ind[y\neq f(\bfx)])
\end{equation}
i.e. the proposed predictor achieves minimum worst-case loss amongst the fair predictors w.r.t. distributions satisfying the two assumptions.
\end{theorem}
\begin{proof}
The proof follows the arguments made in \citep[][Theorem 4]{rojas2018invariant} that proves worst-case optimality of using invariant predictors without considering fairness. For a given source distribution, we will construct an adversarial target distribution which incurs more error for a fair predictor than that of the proposed predictor on the source distribution. This will prove the minmax property from (\ref{eq:worst_case}).

Suppose the density of the source distribution $\mathbb{Q}$ is $q(\bfx,y)$.
Construct a distribution $\mathbb{P}$ with density $p(\bfx,y)=q(\bfs,y)q(\textbf{z})$ where $\textbf{Z}:=\{\textbf{X}\setminus \textbf{S}\}$. 
From the construction of the density, it follows that $\mathbb{P}$ satisfies Assumptions 1 and 2. Thus, $\mathbb{P}\in \mathcal{P}$. 
To see this, observe that $p(y\vert \bfs)=q(y\vert \bfs)$ and $p(\bfs\vert a)=q(\bfs\vert a)$ (for DP, as an example), and we know that $q(y\vert \bfs), q(\bfs\vert a)$ are invariant since $\mathbb{Q}\in \mathcal{P}$.
Also, by construction, $\textbf{Z}\independent (\textbf{S},Y)$ in $\mathbb{P}$.\\

Consider a function $f\in \mathcal{F}(\textbf{G}, \mathcal{P})$. Then,
\begin{align}
\label{eq:worst_case_proof}
\ep_\mathbb{P} (\ind[y\neq f(\bfx)]) \geq \min_{f\in \mathcal{F}(\textbf{G}, \mathcal{P})} \ep_\mathbb{P} (\ind[y\neq f(\bfx)])
\end{align}
Using Lemma \ref{lem:trule}, the error minimizer in $\mathcal{F}(\textbf{G}, \mathcal{P})$ w.r.t. $\mathbb{P}$ is a threshold rule, i.e. $\tilde{f}^\ast(\bfx)=\ind[p_{y\vert \bfx}(\bfx)\geq t_a]$ for a constant $t_a\in [0,1]$. Thus, we can write
\begin{align}
\min_{f\in \mathcal{F}(\textbf{G}, \mathcal{P})} \ep_\mathbb{P} (\ind[y\neq f(\bfx)]) &= \ep_\mathbb{P} (\ind[y\neq \tilde{f}^\ast(\bfx)])\label{eq:worst_case_thresh}\\
&= \ep_\mathbb{P} (\ind[y\neq \tilde{f}^\ast(\bfs,\textbf{z})])\nonumber\\
&= \ep_\mathbb{P} (\ind[y\neq \tilde{f}^\ast(\bfs)])\label{eq:worst_case_proof_1}
\end{align}
where (\ref{eq:worst_case_proof_1}) follows from the construction of $\textbf{Z}$ which satisfies $\textbf{Z}\independent (\bfS,Y)$. Since $p_{y\vert \bfs,\textbf{z}}=p_{y\vert \bfs}$, this implies $\tilde{f}^\ast(\bfs,\textbf{z})=\tilde{f}^\ast(\bfs)$. Plugging back in (\ref{eq:worst_case_proof}), we get
$$\ep_\mathbb{P} (\ind[y\neq f(\bfx)]) \geq \ep_\mathbb{P} (\ind[y\neq \tilde{f}^\ast(\bfs)])$$

By construction, $p(\bfx,y){=}q(\bfs,y)q(\textbf{z})$. Thus, $\ep_\mathbb{P}(g(\bfs,y)){=}\ep_\mathbb{Q}(g(\bfs,y))$ for any function $g$. With this,
\begin{align*}
    \ep_\mathbb{P} (\ind[y\neq f(\bfx)]) &\geq  \ep_\mathbb{Q} (\ind[y\neq \tilde{f}^\ast(\bfs)])\\
    &= \argmin_{f\in \mathcal{F}(\textbf{G}, \mathcal{P})} E_\mathbb{Q}(\ind[y\neq f(\bfs)])
\end{align*}

Thus, for the proposed predictor trained on the source distribution with fairness constraints, there always exists a distribution $\mathbb{P}\in\mathcal{P}$ with larger error for a fair predictor $f\in \mathcal{F}(\textbf{G}, \mathcal{P})$.
In other words, the proposed predictor is worst-case optimal against distributions satisfying the two assumptions.
\end{proof}

\section{Joint Causal Inference (JCI) framework}
\label{app:jci}

JCI framework \citep{mooij2016joint} has been proposed, primarily, for causal discovery from multiple datasets, and for causal modelling of a system observed in different domains (or contexts). Following \citep{magliacane2018domain}, we focus on the latter and use the framework for domain adaptation.

A joint causal model is used to model the data generated under different domains by introducing \textit{context variables} to the standard structural causal model, defined as follows. 
\begin{definition}[Joint Causal Model]
A joint causal model $\mathcal{M}$ is a tuple $\langle\mathcal{I},\mathcal{J},\mathcal{K},\mathcal{H},\mathcal{F},\pd(\mathcal{K})\rangle$ where
\begin{enumerate}
\item $\{C_i\}_{i\in\mathcal{I}}$ denote a set of context variables, assumed to be exogenous to the system of interest.
    \item $\{X_j\}_{j\in\mathcal{J}}$ denote a set of system variables, assumed to be endogenous.
    \item $\{U_k\}_{k\in\mathcal{K}}$ denote a set of independent exogenous variables, assumed to be unobserved.
    \item $\{h_i\}_{i\in\mathcal{H}}$ denote a set of functions that give functional dependence of context variables. Each variable $C_i$ is assigned its value as $C_i \leftarrow h_i(\textbf{U}_{\text{pa}(C_i)\cap \mathcal{K}})$.
    \item $\{f_j\}_{j\in\mathcal{J}}$ denote a set of functions that give functional dependence of system variables. Each variable $X_j$ is assigned its value as $X_j \leftarrow f_j(\textbf{C}_{\text{pa}(X_j)\cap \mathcal{I}}, \textbf{X}_{\text{pa}(X_j)\cap \mathcal{J}}, \textbf{U}_{\text{pa}(X_j)\cap \mathcal{K}})$.
    \item $\pd(\mathcal{K})$ denotes a probability distribution over unobserved exogenous variables, $\pd(\textbf{U}) = \Pi_{k\in\mathcal{K}}\pd(U_k)$.
\end{enumerate}
\end{definition}
Here, $\text{pa}(\cdot)$ is a set of variables, referred to as parents of the variable.

A joint causal graph $\mathcal{G}(\mathcal{M})$ refers to the causal graph representing $\mathcal{M}$. The graph only contains nodes from $\mathcal{I}\cup \mathcal{J}$, a directed edge $a\rightarrow b$ for nodes $a,b$ iff $a\in \text{pa}(b)$, and a bidirected edge $a\leftrightarrow b$ iff the set $\text{pa}(a)\cap\text{pa}(b)\cap\mathcal{K}$ is non-empty. That is, directed edges represent direct functional dependence and bidirected edges represent presence of unobserved common causes.

The functional dependencies along with distribution over the unobserved variables induces a distribution over context and system variables. Importantly, intervening on the context variables give distributions for different domains.\footnote{Since the context variables $\textbf{C}$ are assumed to be exogenous, intervening on them, $do(\textbf{C})$, is same as conditioning on their value, $\textbf{C}=\textbf{c}$. This follows by Rule 2 of do-calculus.} Thus, JCI provides a framework to reason about the distributions of data from multiple domains. To read the conditional independencies in the domain distributions using the graph, we need two assumptions. \textit{Causal Markov assumption} requires that any \textit{d}-separation of sets of nodes in $\mathcal{G}(\mathcal{M})$ imply the corresponding conditional independencies in the distribution. Conversely, \textit{faithfulness} means that there are no other conditional independencies than the ones implied by \textit{d}-separation. Please refer to \citep[][Section 2.2]{magliacane2018domain} for details.

Thus, given any $\mathcal{G}(\mathcal{M})$ proposed using domain knowledge or discovered from data, we can check the conditional independence assumptions posited by Assumptions 1 and 2.

\begin{remark}
In case of TPR and TNR, which require $C\independent S\vert Y=1, A$ and $C\independent S\vert Y=0, A$, we need to consider causal graphs drawn for sub-populations ($Y=1$ or $Y=0$). The conditional independence can then be checked in such a graph using d-separation. This requires stronger assumptions that the modeler can express causal knowledge at the sub-population level and the distribution is faithful to such a graph.
\end{remark}

\section{Results and discussion on counterfactual fairness}
\label{app:counterfactual}

Another set of fairness definitions have been proposed based on the causal effect of the protected attribute on the prediction \citep{kusner2017counterfactual,nabi2018fair,kilbertus2017avoiding}. We consider one strict version of these \textit{counterfactual} fairness definitions.
\theoremstyle{definition}
\begin{definition}{(Ctf) \citep{kusner2017counterfactual}}
A classifier $\hat{Y}=f(\bfX, A)$ is said to be \textit{counterfactually} fair if the interventional distribution of $\hat{Y}$ conditioned on all observed values is the same under $do(A=a)$ and $do(A=d)$, i.e.
$\pd(\hat{Y}_{do(A=a)}=y\vert \bfX=\bfx, A=i)=\pd(\hat{Y}_{do(A=d)}=y\vert \bfX=\bfx, A=i)$,
for $y\in\{0,1\}$ and $i\in\{a,d\}$.
\end{definition}
One method to build a classifier $f(\bfS)$ satisfying Ctf is to only use feature set $\bfS\in\{\bfX,A\}$ that does not contain any descendant of $A$ in the causal graph \citep[][Lemma 1]{kusner2017counterfactual}.

Thus, the counterpart of Assumption 2 for solving Fair DA under Ctf is that the selected feature set contains the non-descendants of $A$. Combined with Assumption 1, we select non-descendants of $A$ which form a separating set in order to solve Fair DA.

Since, Ctf only requires change in feature subset and does not include any fairness constraints in the fair learning problem, we can show the worst-case optimality result as well. The arguments in the proof of Theorem \ref{thm:worst_case_optimal} still hold. The only difference is that instead of using Lemma \ref{lem:trule} at Step (\ref{eq:worst_case_thresh}), we use the fact that Bayes classifier (without fairness constraints) is also a threshold function that depends on conditional outcome distribution.

However, we note that there are multiple ways of defining counterfactual fairness. For instance, \citep{nabi2018fair} require that causal effects of $A$ on $\hat{Y}$ through particular paths should be zero or small.
Further work should explore approaches to solve Fair DA under broader definitions of counterfactual fairness.

\section{Examples of addressable shifts}
\label{app:addressable}

As remarked in Section 5.3, we can find at least one feature set satisfying Assumptions 1 and 2 in case of the following shifts -- shifts with causal paths to $Y$ which all include $A$ (i.e. $C {\cdot\cdot\rightarrow} A{\rightarrow\cdot\cdot} Y$ with all arrows toward $Y$) and shifts with non-causal paths to $Y$ (i.e. $C{\cdot\cdot\rightarrow} M{\leftarrow\cdot\cdot} Y$ for some feature $M\in\bfS$).
This means that any shift causing change in distribution of the sensitive attribute as well as any measurement error in variables with a non-causal path to $Y$ can be addressed.
We present two examples of real-world tasks with such shifts.

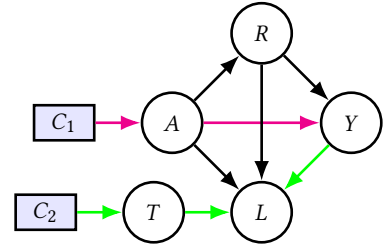
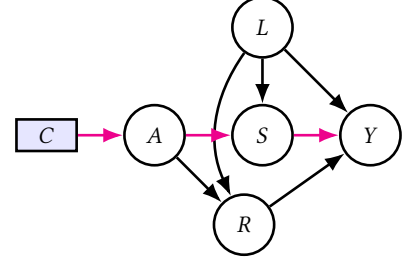
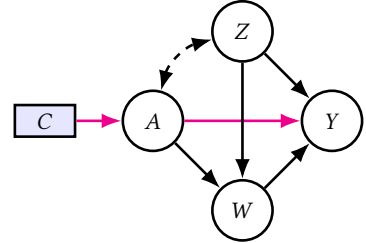
\begin{figure}[t]
    \begin{subfigure}[b]{.4\textwidth}
      \centering
      \begin{tikzpicture}
    \node[state] (a) at (0,0)  {$A$};
    \node[state] (r) [above right =of a] {$R$};
    \node[state] (y) [below right=of r] {$Y$}; 
    \node[state] (l) [below right =of a] {$L$};
    \node[state] (t) [left =of l] {$T$};
     \node[state,rectangle,fill=blue!10] (c2) [left=of t] {$C_2$};
     \node[state, rectangle, fill=blue!10] (c1) [left=of a] {$C_1$};
    \path [color=magenta] (a) edge (y);
    \path (a) edge (r);
    \path [color=green] (y) edge (l);
    \path (r) edge (l);
    \path [color=green] (t) edge (l);
\path [color=green] (c2) edge (t);
    \path [color=magenta] (c1) edge (a);
    \path (a) edge (l);
    \path (r) edge (y);

\end{tikzpicture}       \caption{Disease risk scoring under population shift and treatment policy shift}
      \label{fig:signature_a_app}
    \end{subfigure}\hfill
    \begin{subfigure}[b]{.4\textwidth}
    \centering

\begin{tikzpicture}
    \node[state] (a) at (0,0)  {$A$};
    \node[state] (s) [right =of a] {$S$}; 
    \node[state] (l) [above =of s] {$L$};
    \node[state] (y) [right =of s] {$Y$};
    \node[state] (r) [below right =of a] {$R$};

     \node[state,rectangle,fill=blue!10] (c) [left =of a] {$C$}; 
\path (l) edge (y);
    \path (l) edge (s);
    \path (l) edge[bend right] (r);
    \path [color=magenta] (a) edge (s);
    \path (a) edge (r);
    \path (r) edge (y);
    \path [color=magenta] (s) edge (y);
\path [color=magenta] (c) edge (a);

\end{tikzpicture}     \caption{Credit scoring under population shift}     
    \label{fig:signature_b_app}
    \end{subfigure}\hfill
    \begin{subfigure}[b]{.4\textwidth}
    \centering
    \begin{tikzpicture}
    \node[state] (a) at (0,0)  {$A$};
    \node[state] (w) [below right =of a] {$W$};
    \node[state] (z) [above right =of a] {$Z$};
    \node[state] (y) [below right=of z] {$Y$}; 

     \node[state,rectangle,fill=blue!10] (c) [left =of a] {$C$}; 
    \path [color=magenta] (a) edge (y);
    \path (a) edge (w);
    \path (w) edge (y);
    \path (z) edge (y);
    \path (z) edge (w);
    \path [color=magenta] (c) edge (a);
    \path [bidirected] (a) edge [bend left=30] (z);

\end{tikzpicture}     \caption{Recidivism risk assessment under population shift}     
    \label{fig:signature_c_app}
    \end{subfigure}\caption{Examples of addressable causal graphs. Following Assumptions 1 and 2, including $A$ in the feature set blocks the effect of population shift (e.g. the paths in magenta) and excluding $L$ from the feature set blocks the effect of treatment policy shift (e.g. the path in green).}
\label{fig:signature_app}
\end{figure}

\paragraph{Medical diagnosis}
Figure \ref{fig:signature_a_app} represents an example of sepsis prediction based on the causal graph presented in \cite{subbaswamy2018counterfactual}; $Y$: outcome (prevalence of sepsis condition), $A$: age (sensitive attribute), $R$ represents a pre-existing condition like chronic liver condition. An indicator like the international normalized ratio (INR) which is a measure of blood clotting tendency is denoted by $X$. $T$ represents treatment prescribed to the patient. The graph represents a case of population shift represented by $C_1$ and treatment prescription policy shift denoted by $C_2$. 

\paragraph{Credit scoring}
Figure \ref{fig:signature_b_app} represents a causal graph posited in \citep{chiappa2019path} for the credit scoring task. The task is to predict $Y$: credit risk i.e. whether an applicant will repay the loan or not from features $R$: credit amount and repayment duration, $A$: sex (sensitive attribute), $L$: age, and $S$: savings and housing status. The context variable $C$ denotes an anticipated shift which changes the distribution of females, for instance, in the target domain.

\paragraph{Recidivism risk assessment}
Another example of risk assessment tools used to inform pre-trial bail decisions in the criminal justice system is presented in Figure \ref{fig:signature_c_app}. In the causal graph for this task, presented in \cite{zhang2018equality}, $A$ denotes the sensitive attribute, here race, $Z$ represents other demographic information (e.g., age, gender) of the defendant which can be confounded by race. Prior convictions are denoted by $W$ and recidivism outcome (0 for no, 1 otherwise) is represented by $Y$. $C$ represents population shift where the distribution of the defendant's race changes across jurisdictions i.e. target domains.

\section{Experiment setup}
\label{app:synthsetup}

\textbf{Data generating process for synthetic example.}
We generate data using the following structural equation model.
\begin{align*}
        A &\sim \text{Bernoulli}\left(\sigma\left(\gamma\cdot\lambda_1\cdot C + u_1\right)\right)\\
        R &\sim \cn(0,1) + \lambda_2\cdot A + u_2\\
        Y &\sim \text{Bernoulli}\left(\sigma\left(\lambda_3\cdot A + \lambda_4\cdot R + u_3\right)\right)\\
        T = \lambda_5\cdot &Y + \lambda_6\cdot R + \lambda_7\cdot A + \cn(0,\gamma\cdot\lambda_8\cdot C) + u_4\\
    u_1,u_2,u_3 &\sim \cn(0,0.8^2), u_4\sim \cn(0,1.0)\\
    \lambda_1 &= 0.2, \lambda_2 = -0.1, \lambda_3 = -0.8, \lambda_4 = 0.8\\
    \lambda_5 &= 0.8, \lambda_6 = 0.1, \lambda_7 = -0.8, \lambda_8 = 0.2\\
    \gamma &\in [0,15]
\end{align*}

Variables $A$ and $Y$ are binary, taking values in $\{0,1\}$ and are sampled from a Bernoulli distribution with respective means as per equations, where $\sigma(x)=\frac{1}{1+\text{exp}(-x)}$. For source domain, $C=0$ and for target domain, $C=1$. 
Setting $C=1$ amounts to performing a soft intervention \citep{markowetz2005probabilistic} on $A$ and $T$, whose distributions change as a result in the target domain. The magnitude of the effect of $C$ on $A$ and $T$ is scaled by a constant $\gamma$, which is varied from $0$ to $15$ for simulating distribution shifts of increasing magnitude.

\textbf{More details on $\ot$ baseline.} $\ot$ is an optimal transport-based method for unsupervised domain adaptation \citep{courty2016optimal}. Instead of making assumptions on covariate shift, it posits that some (non-linear) transformation exists between source and target features. It finds the minimal-cost transformations for a given cost function. A classifier is trained on the transformed source data and applied directly to the target data. We use the squared $l_2$ cost and $l_p-l_1$ regularization-based transport method \citep{courty2016optimal}.\footnote{Implemented in POT library \url{https://github.com/rflamary/POT}} Note that we use target features as an input during training for only this baseline.

\section{Synthetic data example: Comparison with Anchor Regression}
\label{app:addexp}

\begin{figure}[htbp!]
\centering
\begin{subfigure}{.5\linewidth}
  \centering
  \includegraphics[scale=0.33]{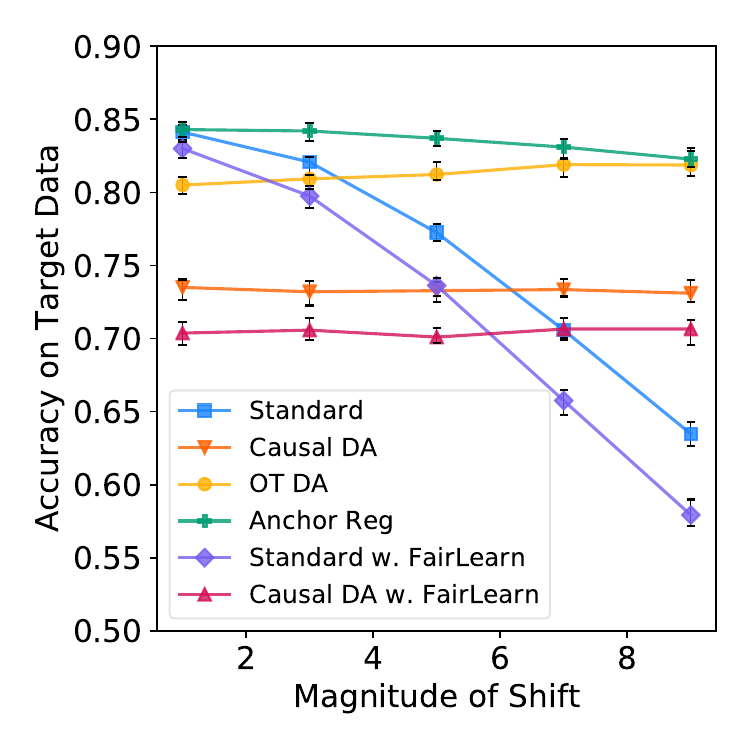}
  \caption{Accuracy vs Shift}
  \label{fig:results_synth_all_acc_ar}
\end{subfigure}\begin{subfigure}{.5\linewidth}
  \centering
  \includegraphics[scale=0.33]{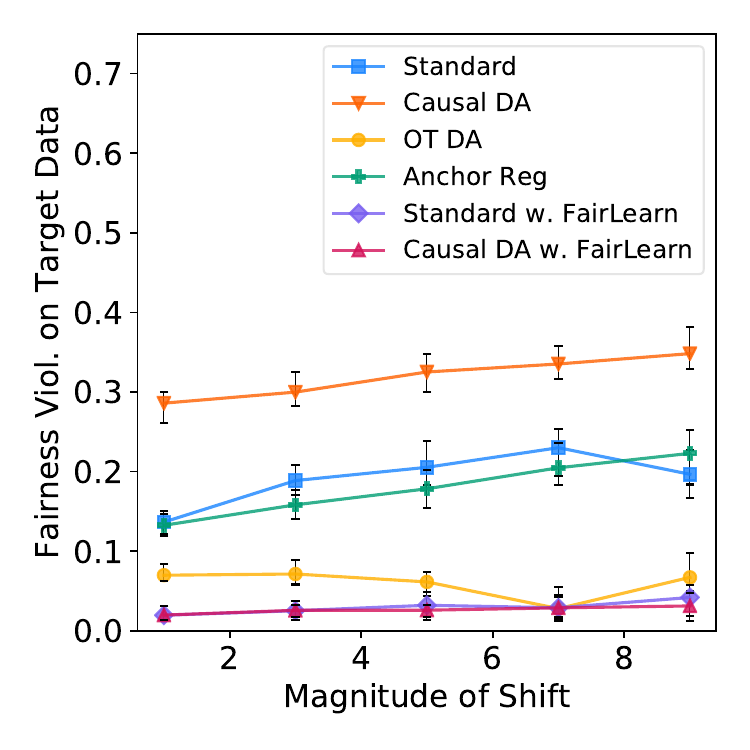}
  \caption{Fairness Violation vs Shift}
  \label{fig:results_synth_all_fair_ar}
\end{subfigure}
\caption{Accuracy and Fairness violation with varying magnitude of shifts in the new setting for synthetic data. Median values are reported over $50$ runs and error bars show first and third quartiles.}
\label{fig:results_synth_all_ar}
\end{figure}

We compare with another causal domain adaptation method called anchor regression \citep{rothenhausler2018anchor} in a separate experiment setting where we assume access to data from multiple source domains. 
This method has been shown to be competitive among state-of-the-art methods for multi-source domain adaptation which do not require target domain data \citep{subbaswamy2019preventing}.

Anchor regression requires specifying an \textit{anchor variable} which indicates (in case of discrete anchors) homogeneous subsets of data. An example is the context variable $C$ which separates the two homogeneous datasets --- source and target --- based on their magnitude of shift.  Anchors are exogenous sources of variation which can be exploited to regularize the predictor. Intuitively, the method avoids overfitting the predictor to the source dataset by de-correlating the residuals from the anchor. In addition, the method assumes
that the anchor causes \textit{mean-shifts} in a subset of variables, i.e. the linear structural equations for the variables have an added intercept term that is constant for a given homogeneous subset. The data generating process described in Section 6.1 does not have an anchor that follows this assumption. The shift in feature $T$ is caused by a shift in the variance of $T$, instead of being a constant mean-shift. Thus, we change the data generating process for the experiments reported here by changing the equation for $T$ (marked in red below).

\begin{align*}
        A &\sim \text{Bernoulli}\left(\sigma\left(\gamma\cdot\lambda_1\cdot C + u_1\right)\right)\\
        R &\sim \cn(0,1) + \lambda_2\cdot A + u_2\\
        Y &\sim \text{Bernoulli}\left(\sigma\left(\lambda_3\cdot A + \lambda_4\cdot R + u_3\right)\right)\\
        T = \lambda_5\cdot &Y + \lambda_6\cdot R + \lambda_7\cdot A + {\color{red} \gamma\cdot\lambda_8\cdot C} + u_4\\
    u_1,u_2,u_3 &\sim \cn(0,0.8^2), u_4\sim \cn(0,1.0)\\
    \lambda_1 &= 0.2, \lambda_2 = -0.1, \lambda_3 = -0.8, \lambda_4 = 0.8\\
    \lambda_5 &= 0.8, \lambda_6 = 0.1, \lambda_7 = -0.8, \lambda_8 = 0.2\\
\gamma &\in [0,9]
\end{align*}

\textbf{Setup} We generate multiple datasets by varying the magnitude of shift, specifically $\gamma=\{0,2,4\}$, and combine them into a single source dataset. In case of anchor regression, dummy variables are added to encode the samples corresponding to the three different shift magnitudes. The dummy variables are the anchors. For rest of the methods, we simply concatenate the datasets without adding any dummy variables. All variables in source and features in target are mean centered using the corresponding datasets. We use square loss as the loss function for the classification problem with an additional regularization term which is minimized following a two-stage least squares procedure \citep{rothenhausler2018anchor}.
We evaluate the resulting predictor on five target datasets $\gamma=\{1,3,5,7,9\}$.

\textbf{Results.} Figure \ref{fig:results_synth_all_ar} shows fairness violation and accuracy for anchor regression along with other methods. We observe that anchor regression (labelled as $\ar$) is robust to distribution shift as it has high accuracy even in case of large shifts (green curve in Figure \ref{fig:results_synth_all_acc_ar}). But it has higher unfairness than the proposed approach $\cdaf$, which is the case for other domain adaptation methods $\cda$ and $\ot$ as well (Figure \ref{fig:results_synth_all_fair_ar}). Thus, this experiment also demonstrates the importance of constraining the fairness violations of the predictors while performing domain adaptation.

\section{Acute Kidney Injury: data description, pre-processing, and additional results}
\label{app:aki}

\subsection{Dataset}
\label{app:data}

The dataset is constructed from a large de-identified electronic healthcare record database, called MIMIC-III, for adult inpatients (Age $>$ 18 years) at a critical care unit of a tertiary care hospital, ranging from years 2001 to 2012 \citep{johnson2016mimic}. The total number of encounters (admissions) in the dataset is $58,976$, with some patients having multiple encounters. In order to have a complete view of a patient's hospital stay from admission until discharge, we removed any patient who died during their hospital stay regardless of the cause. We have also excluded patients that were admitted with evidence of moderate or severe kidney dysfunction, as done by \citet{he2018multi}. Patients with estimated glomerular filtration rate (eGFR) less than $60$ mL/min/$1.73$ $m^{2}$ (calculated using the revised MDRD study equation \citep{levey2007expressing}) or Serum Creatinine (SCr) level more than $1.3$ mg/dL within the first $24$ hours of hospital admission are excluded. Patients admitted with null eGFR or null SCr (i.e., no SCr reading) within first $24$ hours were also removed. The final analysis consisted of $24,852$ encounters.

According to Kidney Disease Improving Global Outcomes (KDIGO)~~\citep{khwaja2012kdigo}, AKI was defined as either of the following two criteria being met: 1) greater than or equal to $50\%$ increase from the baseline SCr value or 2) greater than or equal to $0.3$ mg/dL change in SCr from the baseline creatinine. Baseline SCr level was defined as the first SCr measured after hospital admission. Out of a total of $24,852$ encounters in the final analysis, AKI events occurred in $5,137$ encounters and $19,715$ encounters had no AKI events. We consider a prediction window which ends 1 day before the onset of AKI. A summary of features used to train the prediction models can be found in Table~\ref{tab:features}, included at the end. Some of these are also used by \cite{he2018multi}. However, unlike this study, we exclude medications and medical history features as they are high-dimensional (around 1,000). Only patients who did not have AKI at the time of admission are included since we are interested in diagnosing AKI for in-hospital patients. Data for variables mentioned in Table~\ref{tab:features} is extracted for each encounter from time before the prediction window.
Since the features are measured at multiple timepoints, we consider the last observed value of a feature until the end of the prediction window.

\subsection{Data pre-processing}
\label{app:akipreprocess}
In-hospital encounters which met the two KDIGO criteria were labeled as the positive class, while encounters in which the patient did not meet the above criteria was labeled as the negative class. Feature vectors were created for each encounter. Demographic information (age, gender, and race) were included in the feature vectors; age in years, and dummy variables for gender categories (female, male) and race categories (American Indian or Alaska Native, Asian, Black or African American, Native Hawaiian or Other Pacific Islander, White, multiple race, refuse to answer, no information, unknown, and other). 
A patient's vitals (BMI, diastolic BP, systolic BP, height, weight) were included as numerical features, where the most recent value associated with any vital was used in cases where multiple measurements were taken during an encounter. Any missing values were imputed using the mean value (calculated on the training data) for the given feature. Lab tests and comorbidities were included in the form of boolean features indicating whether the lab test/comorbidity was present. While only lab tests performed during a given in-hospital encounter were used, comorbidities up to one year prior to the hospital stay were included. Comorbidities included in the AKI predictive model by \cite{tomavsev2019clinically} were used as a guide for selecting the nine comorbidities in Table~\ref{tab:features}. After adding dummy variables and missing value indicators, we obtained a dataset containing $24,852$ examples, each with $51$ features ($13$ demographic, $5$ vitals, $9$ comorbidities, $12$ lab tests, and $12$ lab test missing value indicators). We removed features that had more than $90\%$ values as missing, namely the lab tests for Troponin, Albumin, and WBC. 
Since the BUN lab test was one of the most predictive features, it was chosen for creating shifted datasets. We discarded both the BUN variable and missing value indicator when creating the invariant predictor.

\begin{figure*}[t]
\centering
\begin{subfigure}{.33\linewidth}
  \centering
  \includegraphics[scale=0.33]{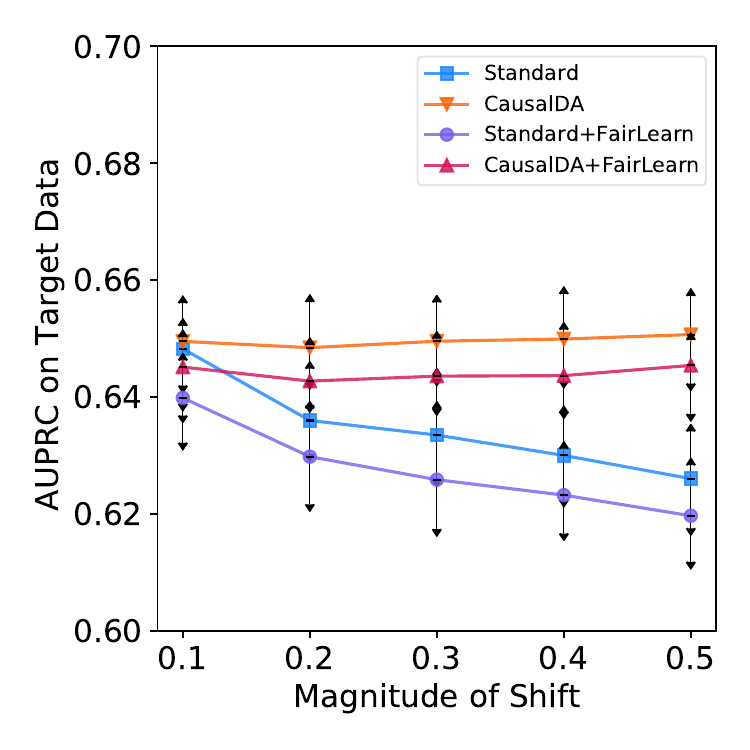}
  \caption{AUPRC vs Shift.}
  \label{fig:results_mimic_auprc_all}
\end{subfigure}\begin{subfigure}{.33\linewidth}
  \centering
  \includegraphics[scale=0.33]{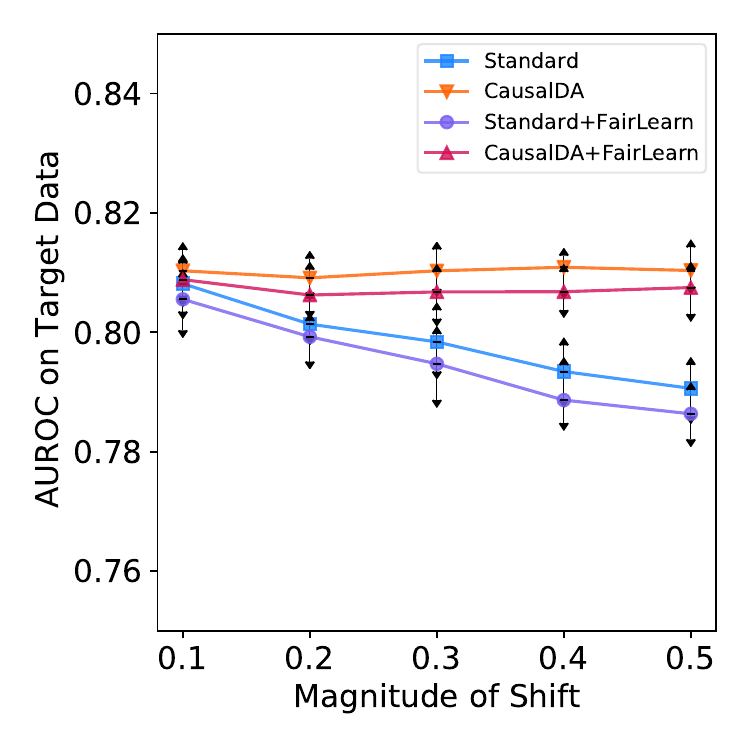}
  \caption{AUROC vs Shift.}
  \label{fig:results_mimic_roc_auc_all}
\end{subfigure}\begin{subfigure}{.33\linewidth}
  \centering
  \includegraphics[scale=0.33]{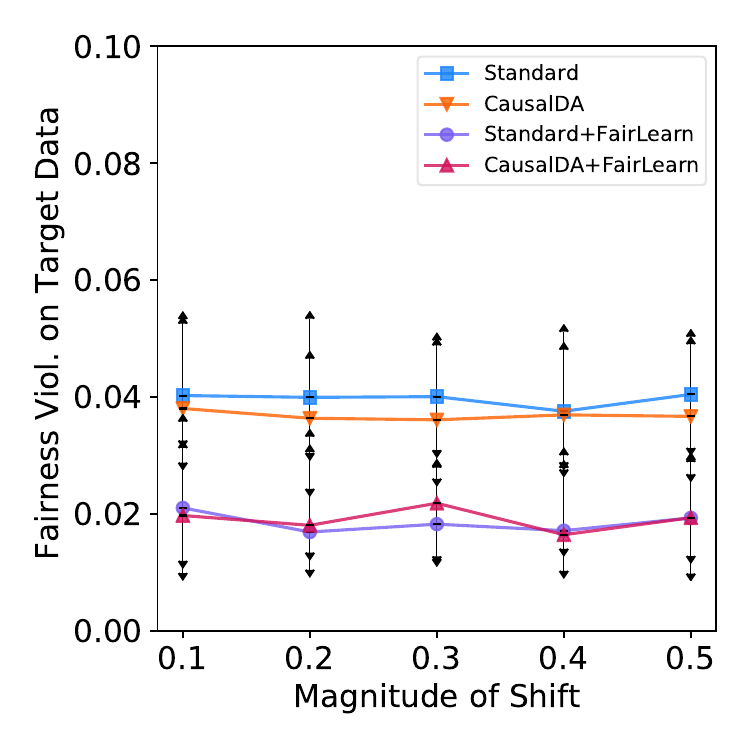}
  \caption{Fairness Violation vs Shift.}
  \label{fig:results_mimic_fair_all}
\end{subfigure}
\caption{Accuracy, AUROC, and Fairness violation with varying magnitude of shifts for the AKI task. Median values are reported over $50$ runs and error bars show first and third quartiles.}
\label{fig:results_mimic_all}
\end{figure*}

\begin{remark}
We include race categories, constructed from race and ethnicity codes in the raw data, as a feature in the model following past work on AKI risk assessment \citep{he2018multi, zimmerman2019early}. However, we emphasize that race is an ill-defined and ill-measured feature and including it might cause discrimination in allocating healthcare resources based on the model predictions \citep{eneanya2019race, vyas2020race}. Thus, more careful examination of the decision to include race in the models is required.
\end{remark}

\subsection{Different proportions of missingness}
\label{app:propmissing}
To evaluate robustness of models, distributional shifts are simulated in target data following the setup outlined in Section 6.2. Firstly, we add missing values in the BUN feature values for a proportion of the rows. Secondly, we downsample the female population by randomly keeping only $50\%$ of the rows from the female group in the source data. 
Figure \ref{fig:results_mimic_all} shows two predictive performance metrics, namely AUPRC, AUROC, and one fairness metric, namely maximum fairness violation, for different proportions of missingness, ranging from $0.1$ to $0.5$. As mentioned in Section 6.2, the methods which use the invariant features ($\cda$ and $\cdaf$) have stable accuracy as the distribution shifts. Moreover, $\cdaf$ reduces unfairness consistently in the shifted data. 
Table \ref{tab:results_mimic} reports the numerical values for evaluation metrics with missingness proportion set to $0.5$. These are reported visually as well in Figure \ref{fig:results_mimic_all} here and Figure 4c in the main text.

\begin{table}[htbp!]
    \centering
    \caption{Fairness Violation, AUPRC and AUROC (median across 50 runs) for AKI task, with missigness proportion set to 0.5}
    \label{tab:results_mimic}
    \begin{tabular}{|c|c|c|c|}
        \hline
        Model & Fairness Viol. & AUPRC & AUROC \\
        \hline
        $\std$ & 0.0404 & 0.626 & 0.7906 \\
        \hline
        $\cda$ &  0.037 & 0.6507 & 0.8104\\ 
        \hline
        \specialcell{$\stdf$} & 0.0193 & 0.6197 & 0.7863 \\
        \hline
        \specialcell{$\cdaf$} & 0.0193 & 0.6454 & 0.8075 \\
        \hline
    \end{tabular}
\end{table}

\begin{table}[htbp!]
    \centering
    \caption{Hyperparameter settings used in synthetic and AKI experiments}
    \label{tab:hyperparam}
    \begin{tabular}{|c|c|c|}
        \hline
        Method & Hyperparameter & Value \\
        \hline
\multirow{3}{*}{\specialcell{$\fl$ -\\ Exponentiated\\ Gradient}} & Allowed fairness constraint violation (eps) & $10^{-2}$ \\
        & Maximum number of iterations (T) & 50 \\
        & Convergence threshold (nu) & $10^{-6}$ \\
        \hline
         \multirow{2}{*}{$\ot$} & Entropic regularization (reg\_e) & 10 \\
         & Class regularization (reg\_cl) & $10^{-2}$ \\
        \hline
        $\ar$ & Regularization coefficient (gamma) & 1.5 \\
        \hline
    \end{tabular}
\end{table}

\subsection{Implementation details}
\label{app:hyperparam}
Hyperparameters used in solving the Fair DA problem are listed in Table \ref{tab:hyperparam}. Rest of the hyperparameters are kept as defaults chosen in the packages cited. Classifiers (logistic regression and gradient boosting trees) are implemented using the package \texttt{scikit-learn} (v0.22.1) \citep{scikit-learn} in Python.
All experiments were ran on a single node of a compute cluster with a 3.0 GHz Intel processor and 8 GB memory.

\begin{table*}[t]
\centering
\caption {Covariates for Training AKI Model on MIMIC-III Data.}
\label{tab:features}
\small{
\begin{tabular}{lcp{0.8\textwidth}}
\hline
\thead{Feature \\ Category} & \thead{Number \\ of Variables} & \thead[l]{Details} \\
\hline
Demographics & 3 & \vspace{-2mm}\begin{itemize}[leftmargin=1em,topsep=0em]
    \tightlist
    \item Age
    \item Gender
    \item Race
\end{itemize} \\
Vitals & 5 & \vspace{-2mm}\begin{itemize}[leftmargin=1em,topsep=0em]
    \tightlist
    \item BMI
    \item Diastolic BP
    \item Systolic BP
    \item Height
    \item Weight
\end{itemize} \\
Lab tests & 12 & \vspace{-2mm}\begin{itemize}[leftmargin=1em,topsep=0em]
    \tightlist
    \item Albumin, Body Fluid
    \item Alanine Aminotransferase (ALT)
    \item Asparate Aminotransferase (AST)
\item Bilirubin, Total
    \item Blood Urea Nitrogen (BUN)
    \item Free Calcium
\item Creatine Kinase (CK)
    \item Glucose
    \item Lipase
    \item Platelets
    \item Troponin I
    \item WBC, CSF
\end{itemize} \\
Comorbidities & 9 & \vspace{-2mm}\begin{itemize}[leftmargin=1em,topsep=0em]
    \tightlist
    \item Diabetes Mellitus w/ Complications
    \item Diabetes Mellitus w/o Complication
    
    \item Gout \& Other Crystal Arthropathies
    \item Hypertension w/ Complications \& Secondary Hypertension
    \item Chronic Obstructive Pulmonary Disease \& Bronchiectasis
    \item Chronic Kidney Disease
    \item Hypertension Complicating Pregnancy; Childbirth \& the Puerperium
    \item Diabetes or Abnormal Glucose Tolerance Complicating Pregnancy; Childbirth; or the Puerperium
    \item Chronic Ulcer of Skin
\end{itemize}\\
\hline
\end{tabular}
}
\end{table*}

\end{document}